\documentclass[10pt,twocolumn,letterpaper]{article}

\usepackage[pagenumbers]{cvpr} 

\definecolor{cvprblue}{rgb}{0.21,0.49,0.74}
\usepackage[pagebackref,breaklinks,colorlinks,allcolors=cvprblue]{hyperref}

\usepackage{siunitx}
\usepackage{multicol}
\usepackage{graphicx}
\usepackage{booktabs}

\usepackage{multirow}
\usepackage[dvipsnames,table]{xcolor}
\usepackage{colortbl}

\usepackage{multirow}

\definecolor{factor1}{RGB}{11, 94, 163}
\definecolor{factor2}{RGB}{212, 124, 9}
\definecolor{factor3}{RGB}{19, 143, 96}
\definecolor{factor4}{RGB}{201, 73, 6}

\def\paperID{15077} 
\def\confName{CVPR}
\def\confYear{2025}

\title{On the Generalization of Handwritten Text Recognition Models}

\author{Carlos Garrido-Munoz \quad Jorge Calvo-Zaragoza\\
Pattern Recognition and Artificial Intelligence Group, University of Alicante, Spain\\
{\tt\small \{carlos.garrido,jorge.calvo\}@ua.es}
}
\graphicspath {{figures/}}
\begin{document}
\maketitle
\begin{abstract}
Recent advances in Handwritten Text Recognition (HTR) have led to significant reductions in transcription errors on standard benchmarks under the i.i.d. assumption, thus focusing on minimizing in-distribution (ID) errors.
However, this assumption does not hold in real-world applications, which has motivated HTR research to explore Transfer Learning and Domain Adaptation techniques. In this work, we investigate the unaddressed limitations of HTR models in generalizing to out-of-distribution (OOD) data. 
We adopt the challenging setting of Domain Generalization, where models are expected to generalize to OOD data without any prior access. To this end, we analyze 336 OOD cases from eight state-of-the-art HTR models across seven widely used datasets, spanning five languages. Additionally, we study how HTR models leverage synthetic data to generalize. We reveal that the most significant factor for generalization lies in the textual divergence between domains, followed by visual divergence. We demonstrate that the error of HTR models in OOD scenarios can be reliably estimated, with discrepancies falling below 10 points in 70\% of cases. 
We identify the underlying limitations of HTR models, laying the foundation for future research to address this challenge.
Code is available at \href{https://www.github.com/carlos10garrido/HTR-OOD}{github.com/carlos10garrido/HTR-OOD}. 
\end{abstract}

\section{Introduction}
\label{sec:intro}

\begin{figure}[!ht]
    \centering
    \includegraphics[width=1.0\linewidth]{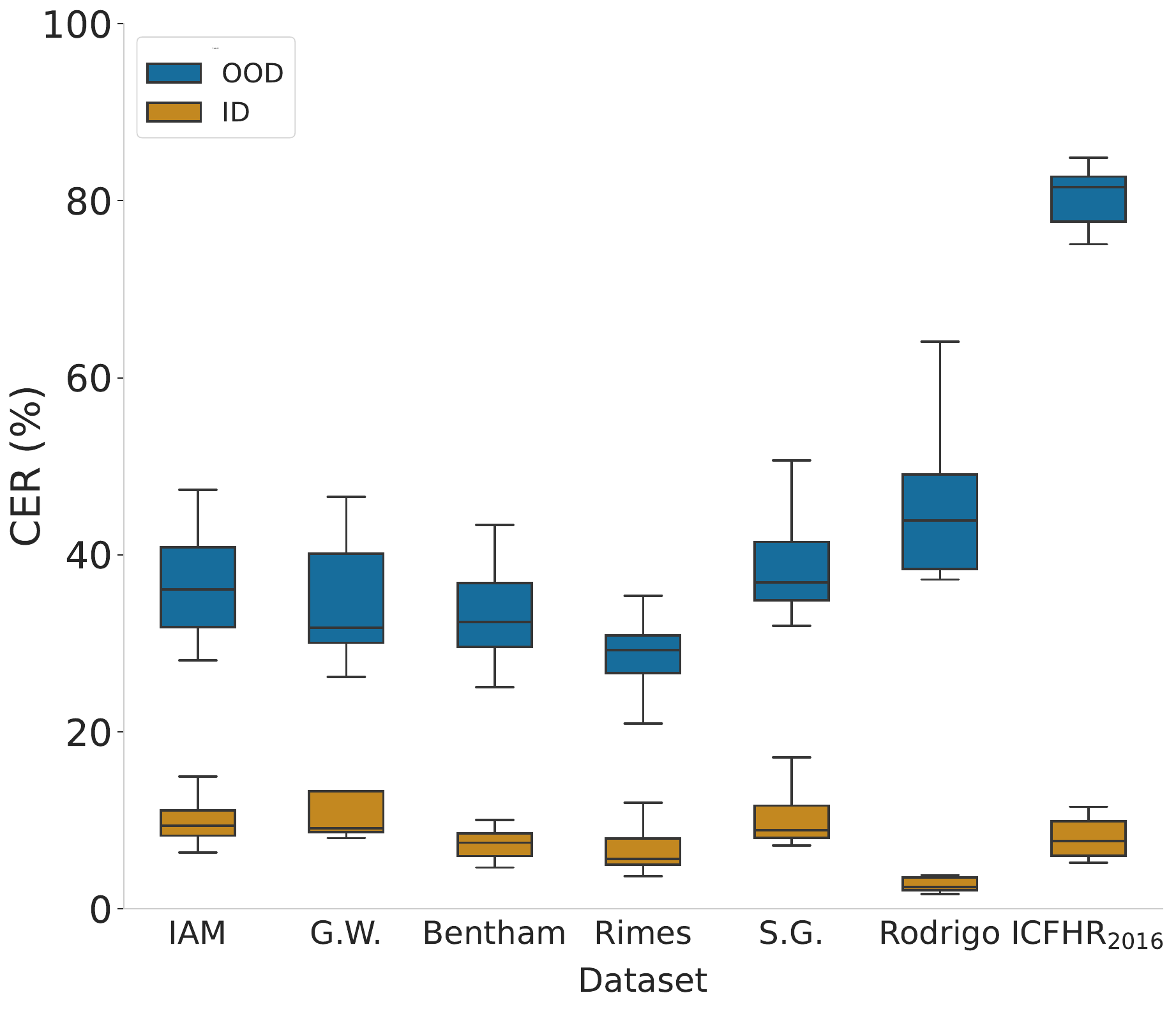}
    \caption{Average performance of HTR models for in-distribution (ID) and out-of-distribution (OOD) scenarios. The significant performance drop in OOD scenarios highlights the limited generalization capability of current models. Details in Sect. \ref{sec:practical_analysis}.}
    \label{fig:id-ood}
\end{figure}

 \begin{figure*}[ht!]
    \centering
    \includegraphics[width=1.0\linewidth]{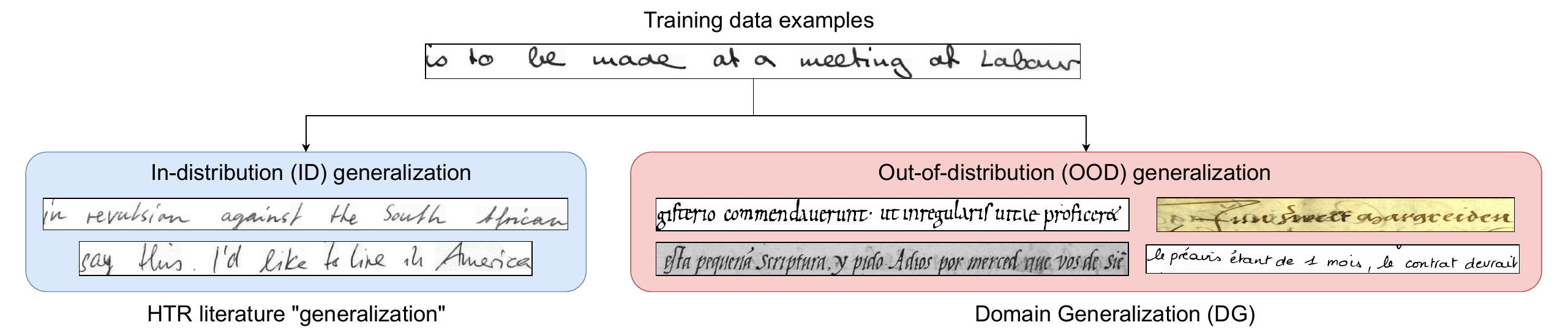}
    \caption{Illustration of the different approaches for the concept of ``generalization'' in HTR literature. While the field commonly considers in-distribution (ID) train--test scenarios as 'generalization' (variations within similar scripts and languages), true out-of-distribution (OOD) generalization---encompassing vastly different scripts, languages, and historical periods---is not addressed.}
    \label{fig:dg_setting}
\end{figure*}

Handwritten text plays a central role in transmitting human knowledge and culture across generations. Despite the advancements in digital storage, significant volumes of handwritten documents remain to be transcribed and thus inaccessible for further analysis or retrieval. The field of Handwritten Text Recognition (HTR) focuses on automatically transcribing handwritten text into digital format. HTR has wide-range applications, including the preservation of cultural heritage \cite{iamdatabase_marti_2002}, digital transcription of historical and matrimonial records \cite{esposalles_2013}, digitization of personal and official documents \cite{washington_2007}, and even real-time digital note-taking through graphic tablets \cite{read_htr_tablet_2007}. However, this transcription process is challenging due to the vast variations in writing styles, different languages and alphabets, and even the mediums on which the text is written~\cite{scalable_htr_ingle_2019}.


Perhaps influenced by the long-standing competitions in the field of HTR \cite{boosting_aradillas_2021, icdar2017_snchez_2017, icfhr2014_snchez_2014, international_abed_2010}, the latest improvements---from both models \cite{multidimensional_recurrent_puig_2017, endtoend_coquenet_2022, pay_attention_kang_2022, light_barrere_2022, training_barrere_2024, li_htr-vt_2025, trocr_li_2023} and data generation \cite{generating_krishnan_2016, handwritten_pippi_2023, synthetic_jaderberg_2014, vatr_vanherle_2024, bhunia_handwritting_transformers_2021}---have been oriented towards increasing performance in the test sets of standard benchmarks, relying on the assumption that test data will conform to the same source distribution. This adheres to the independent and identically distributed (i.i.d) condition, which is often overlooked in real-world settings \cite{domino_eyuboglu_2022, recognition_beery_2018, theory_bendavid_2010}. However, as demonstrated by our preliminary results shown in Fig.~\ref{fig:id-ood}, the performance gap between in-distribution (ID) data and out-of-distribution (OOD) data in HTR is particularly remarkable. This phenomenon is yet to be explored in this field. Although this generalization issue remains largely unexplored, the literature recognizes the importance of generalizing beyond the training data and focuses on two main approaches: (1) Transfer Learning (TL), where pre-trained models are fine-tuned on a target domain \cite{survey_weiss_2016, comprehensive_zhuang_2019, survey_tan_2018}, and (2) Unsupervised Domain Adaptation (UDA), as a particular case of TL, that aims to adapt prior knowledge to new unlabeled data \cite{deep_wang_2018, survey_wilson_2018} in synth-to-real scenarios \cite{kang_unsupervised_2020} or with real data \cite{bhunia_metahtr_2021, soullard_improving_2019}.
However, TL approaches in HTR are constrained to scenarios where the source and target domains share similar characteristics \cite{aradillas_jaramillo_boosting_2018, kohut_towards_2023} or limited to a single architecture \cite{boosting_aradillas_2021}. Likewise, existing UDA studies are limited by small datasets \cite{soullard_improving_2019, zhang_sequence--sequence_2019} or focus on word-level tasks \cite{kang_unsupervised_2020, bhunia_metahtr_2021}, restricting their applicability to general HTR at the line level. In this work, departing from the TL and UDA literature, we rather adopt a Domain Generalization (DG) framework, which aims to generalize to OOD data---data significantly different from the training set---without any prior access to it \cite{domain_zhou_2021, generalizing_wang_2021}.

Given the scarcity and diversity of labeled datasets in the HTR field \cite{scalable_htr_ingle_2019}, we stress on a single-domain DG formulation \cite{learning_wang_2021, learning_qiao_2020} moving away from the deceiving concept of ``generalization'' in the HTR literature (see Fig. \ref{fig:dg_setting}). Our experimentation comprises two approaches. First, we provide practical insights into the generalization capabilities of HTR models, addressing questions such as: What is the best model in terms of generalization? Is there a general HTR method that outperforms others in generalization? In this first part, we also explore how key factors from the DG literature, such as model selection \cite{in_search_lost_dg_2021} and model capacity \cite{gouk2024limitationsgeneralpurposedomain} enhance generalization performance. Additionally, we explore which models best leverage synthetic data to predict on real sets. The second part focuses on identifying the factors that most impact generalization in HTR models. To this end, we conduct a factor analysis \cite{general_spearman_1904, confirmatory_wood_2008, frontend_dehak_2011} to reveal the most significant contributors. Before determining these factors, we address two critical challenges in HTR: (1) quantifying visual divergence between domains, which refers to measuring how different the visual features of characters or sentences are across various domains and (2) assessing textual disparities, which involves evaluating differences in the underlying linguistic content.  These two challenges are addressed with two metrics---visual divergence and textual divergence---that intuitively should play a significant role in explaining OOD performance. Finally, leveraging the identified factors, we assess whether it is possible to estimate the amount of generalization error in advance, following the OOD literature on robustness \cite{accuracy_miller_2021,agreementontheline_baek_2022}. To mitigate potential biases in our conclusions, we analyze 336 distinct OOD evaluations under consistent and standardized conditions, examining 8 distinct HTR models from the literature across 7 datasets covering 5 different languages: English, Spanish, French, German, and Latin. 

Our key contributions are: (1) We conduct the first large-scale analysis of HTR model performance evaluating both ID and OOD scenarios and provide the first comprehensive cross-lingual generalization study using real and synthetic datasets; (2) We introduce and analyze proxy metrics for explaining HTR performance in OOD scenarios; 
(3) We identify key factors that most influence generalization in HTR; (4) We demonstrate that OOD error can be estimated robustly from the considered proxies; (5) We provide a thorough generalization analysis framework in HTR that provides the groundwork for future research in the field.


\section{Related Work}
\label{sec:related-work}

\subsection{Handwritten Text Recognition}
The use of bidirectional Long Short-Term Memory (LSTMs) networks \cite{Bi-LSTMGRAVES2005602, Hochreiter-LSTM} with Connectionist Temporal Classification (CTC)~\cite{connectionist_graves_2006} has been dominating the state of the art in HTR competitions for decades \cite{boosting_aradillas_2021, icdar2017_snchez_2017, icfhr2014_snchez_2014, international_abed_2010}. Despite this prevalence of CTC, attention-based encoder-decoder approaches \cite{Bahdanau:ICLR:2015} have recently gained popularity because of their competitive results \cite{attentionhtr_kass_2022, lexicon_kumari_2022, attentionbased_abdallah_2020, endtoend_coquenet_2022}. The work of Michael et al. \cite{evaluating_michael_2019} provides a comprehensive study of several sequence-to-sequence approaches for HTR. 
There has also been a growing interest in the use of more scalable and parallelizable architectures such as the Transformer \cite{TransformerVaswani} by adapting the work of \cite{Dosovitskiy2020AnII} (Vision Transformer) to the field \cite{training_barrere_2024, weakly_paul_2023, improving_sang_2019, dtrocr_fujitake_2023, rethinking_diaz_2021, characterbased_poulos_2021}. HTR has benefited from this adaptation, either in isolation with an encoder-decoder \cite{trocr_li_2023, transformerbased_momeni_2023, ocformer_mostafa_2021, transformer_wick_2021} or in combination with the CTC objective function \cite{rescoring_wick_2021, dan_coquenet_2023, light_barrere_2022}. The work of Diaz et al. \cite{rethinking_diaz_2021} explores universal architectures for text recognition, concluding that a CNN backbone with a Transformer encoder, a CTC-based decoder, plus an explicit language model, is the most effective strategy to date. Regardless of this progress, however, the need for large labeled corpora as a pre-training strategy in Transformer-based models has become noticeable \cite{trocr_li_2023, rethinking_diaz_2021, vilbert_lu_2019, xlnet_yang_2019, simpler_lu_2021}. This issue could be mitigated through the use of Self-Supervised Learning (SSL) training methods \cite{reading_yang_2022, selfsupervised_pearrubia_2024, sequencetosequence_aberdam_2021}.

\subsection{Adaptation and Generalization in HTR}
\paragraph{Transfer Learning (TL).}
Transfer Learning \cite{survey_weiss_2016, comprehensive_zhuang_2019, survey_tan_2018} has emerged as a popular approach to improve HTR systems, particularly when dealing with small datasets or adapting models to new domains. \cite{aradillas_jaramillo_boosting_2018} investigated the knowledge transfer from larger datasets to smaller ones, focusing on which layers of neural networks require retraining. In a follow-up study, \cite{aradillas_boosting_2021} further explored the combination of Domain Adaptation (DA) and TL, concluding that while it yields the best results, TL alone can achieve nearly comparable performance. Aligned with these findings, \cite{kohut_towards_2023} demonstrated that fine-tuning is surprisingly effective as a domain adaptation baseline in handwriting recognition, focusing on architectures using CTC. \cite{pippi_how_2023} explored pre-training strategies for HTR models, including the use of synthetic data and data generated by Handwritten Text Generation (HTG) \cite{handwritten_pippi_2023, vatr_vanherle_2024}, evaluating scenarios where only the target language or author information is known. 

\paragraph{Domain Adaptation (DA).} 
The closest field to our research is that of Unsupervised Domain Adaptation (UDA) \cite{deep_wang_2018, survey_wilson_2018}, sometimes referred to as Writer Adaptation (WA) in the HTR literature, which investigates techniques with which to adapt HTR models trained on samples from either one or multiple writers to unseen writers \cite{zhang_sequence--sequence_2019, soullard_improving_2019, domain_zhou_2021, van_der_werff_writer_2023, tula_is_2023, kohut_towards_2023} or adapting synth-to-real data \cite{kang_unsupervised_2020}. This field explores adaptation to new data by assuming access to unlabeled samples from the target domain. In contrast, our work is more akin to that of Domain Generalization (DG) \cite{domain_zhou_2021, in_search_lost_dg_2021, generalizing_wang_2021, learning_wang_2021, learning_qiao_2020}, where there is no access to any type of data from the OOD target. To the best of our knowledge, this scenario has not been explored in HTR. Nonetheless, \cite{pippi_how_2023} introduces some close-to-DG experiments, focusing on examining TL from several real-world source datasets to only a few real-world target datasets, without deeply exploring the generalization capabilities of these models. Instead, authors use this scenario to compare various strategies using different percentages of target data, blending both synthetic and real datasets within a TL framework. In our work, we rather focus on studying the generalization capabilities of HTR models to entirely new manuscripts without making assumptions about the availability of specific unlabeled data from the target domain. This approach is more challenging (as depicted in Fig.~\ref{fig:dg_setting}), since the robustness to the OOD scenario has to be performed without prior information about the new domain. We emphasize that in this work we do not apply any type of adaptation to any target dataset, in contrast to the WA/DA field.
\section{Methodology}
\label{sec:methodology}

\subsection{HTR formulation}
HTR models take an image $\mathbf{x}$ and predict a sequence $y = (y_1, y_2, \dots y_L)$, where each $y_i$ is a character from an alphabet $\Sigma$. 
Modern HTR models are trained in an end-to-end fashion to align image-text pairs $(\mathbf{x}, \mathbf{y})$ from a training set $\mathcal{D_\text{train}}$. Then, they are evaluated on a test set $\mathcal{D_\text{test}}$ to measure ``generalization'' performance. In this paper, we explore the generalization capabilities of HTR models in unseen domains, addressing a more challenging task than previously reported in the literature and aligning more closely with the DG paradigm, as illustrated in Fig. \ref{fig:dg_setting}. 
Note that this type of generalization extends beyond the typical author-to-author generalization in HTR \cite{soullard_improving_2019, bhunia_metahtr_2021,bhunia_text_2021} and addresses new OOD challenges such as new manuscripts, alphabets, and languages. 

\subsection{Baseline HTR Models}
We have selected eight distinct architectures \cite{multidimensional_recurrent_puig_2017, endtoend_coquenet_2022, arce_self-attention_2022, li_htr-vt_2025, pay_kang_2020, evaluating_michael_2019, light_barrere_2022, training_barrere_2024} specifically designed for HTR, representing all this broad spectrum. Despite the number of methods, these can be primarily organized by their approach to align the input image $x$ with the output sequence $y$, fitting into one of the three main categories: (1) Connectionist Temporal Classification (CTC) \cite{multidimensional_recurrent_puig_2017,endtoend_coquenet_2022,arce_self-attention_2022,li_htr-vt_2025}, (2) Sequence-to-Sequence models \cite{pay_kang_2020}, or (3) Hybrid models \cite{evaluating_michael_2019, light_barrere_2022, training_barrere_2024}. 

\paragraph{CTC.}
Most architectures in the modern HTR literature employ the CTC objective \cite{connectionist_graves_2006} as an effective method for aligning images with unsegmented transcriptions. We selected four distinct architectures with different feature extractors. These architectures are: CRNN \cite{multidimensional_recurrent_puig_2017}, VAN \cite{endtoend_coquenet_2022}, CNN-SAN \cite{arce_self-attention_2022}, HTR-VIT \cite{li_htr-vt_2025}. This selection comprises different CTC-based architectures with a broad range of encoders and decoders.

\paragraph{Sequence-to-sequence.}
Sequence-to-sequence models (Seq2Seq) have been dominating the arena of language processing~\cite{Bahdanau:ICLR:2015}, particularly boosted by the introduction of the Transformer \cite{TransformerVaswani}.\footnote{
In this work, we do not include TrOCR \cite{trocr_li_2023}, as it was not originally designed for HTR but rather as a secondary application.} To investigate this type of Seq2Seq architectures, we utilize the Transformer presented in \cite{pay_kang_2020} (ResNet + Transformer) since other pure Transformer-based architectures merely differ in post-processing techniques \cite{transformer_wick_2021}. 

\paragraph{Hybrid.}
Several studies have adopted hybrid approaches for HTR \cite{evaluating_michael_2019, light_barrere_2022, training_barrere_2024}. Inspired by \cite{joint_hybrid_speech_kim_2017}, which combined CTC for the encoder and Cross-Entropy (CE) for the decoder as a multitask learning for Speech Recognition, a similar method was first applied in HTR in \cite{evaluating_michael_2019}. This approach utilizes a hybrid loss function that merges CTC and CE losses as $\mathcal{L} = \lambda\mathcal{L_{\text{ctc}}} + (1 - \lambda)\mathcal{L_{\text{ce}}}, \lambda \in [0, 1]$,

with $\lambda$ typically set to 0.5. This combined loss has proven beneficial in HTR, as shown in \cite{evaluating_michael_2019}. Additionally, recent trends in HTR explore transcription as a multitask learning problem, particularly with lightweight CNN and Transformer architectures \cite{light_barrere_2022, training_barrere_2024}. We explore this late trend in HTR throughout the architectures proposed in \cite{evaluating_michael_2019, light_barrere_2022, training_barrere_2024}. 

Table \ref{tab:summary_architectures} summarizes the architectures considered in this work.

\begin{table}[htbp]
\centering
\caption{Description of HTR architectures considered, including alignment type, number of parameters, and input image sizes. \small C=CNN; FCN=Fully Convolutional Network; SA=Self Attention; T=Transformer; Att=``Classic'' attention.
}
\small 
\begin{tabular}{@{}p{1.9cm}p{1.7cm}p{0.5cm}rr}
\toprule
\textbf{Model} & \textbf{Architecture} & \textbf{Align.} & \textbf{Params} & \textbf{Input size} \\
\midrule
\small CRNN \cite{multidimensional_recurrent_puig_2017}   & CRNN         & CTC      & 9.6M  & 128,1024 \\ 
\small VAN \cite{endtoend_coquenet_2022}    & FCN          & CTC      & 2.7M  & 64,1024 \\ 
\small C-SAN \cite{arce_self-attention_2022}        & C+SA        & CTC      & 1.7M  & 128,1024 \\ 
\small HTR-VT \cite{li_htr-vt_2025}       & C+ViT       & CTC      & 53.5M & 64, 512 \\ 
\midrule
\small Kang \cite{pay_attention_kang_2022}         & ResNet+T   & Seq2Seq  & 90M   & 64,2227 \\ 
\midrule
\small Michael \cite{evaluating_michael_2019}      & CRNN+Att.Dec & Hybrid   & 5M    & 64,1024 \\ 
\small LT \cite{light_barrere_2022}  & C+T.Enc+CTC & Hybrid & 7.7M  & 128,1024 \\ 
\small VLT  \cite{training_barrere_2024} & C+T.Enc+CTC & Hybrid & 5.6M  & 128,1024 \\ 
\bottomrule
\end{tabular}
\label{tab:summary_architectures}
\end{table}

\subsection{Experimental setup}
\paragraph{Data.} We assessed the performance of the models in the following datasets: IAM \cite{iamdatabase_marti_2002}, Rimes \cite{rimes_2010}, Bentham \cite{bentham_causer2012building}, Saint-Gall (S.G.) \cite{saint_gall_scherrer1875verzeichniss}, George Washington (G.W.) \cite{washington_2007}, Rodrigo \cite{serrano-etal-2010-rodrigo} and ICFHR 2016 (READ 2016) \cite{icfhr_2016_sanchez}. Table \ref{tab:dataset-comparison} details each dataset. 

\begin{table}[h]
\centering
\caption{Handwritten document datasets considered, with an indication of the language, historical period, number of writers, number of samples in each partition, and size of the alphabet ($\Sigma$).} 
\small
\begin{tabular}{@{}p{1.7cm}p{0.35cm}cp{0.3cm}p{0.35cm}p{0.35cm}p{0.35cm}c@{}}
\toprule
\textbf{Dataset} & \textbf{Lang.} & \textbf{Period} & \textbf{No. writ} & \textbf{Train} & \textbf{Val} & \textbf{Test} & \textbf{|$\Sigma$|} \\ 
\midrule
IAM & En & 1999 & 657 & 6.4K & 976 & 2.9K & 79 \\
Rimes & Fr & 2011 & 1.3K & 10K & 1.1K & 778 & 100 \\
Bentham & En & 18-19th c. & 1 & 9.1K & 1.4K & 860 & 91 \\
S.G.  & Lat & 9-12th c. & 1 & 1.4K & 235 & 707 & 49 \\
G.W. & En & 1755 & 1 & 325 & 168 & 163 & 68 \\
Rodrigo & Sp & 1545 & 1 & 20K & 1K & 5K & 115 \\
ICFHR$_{2016}$ & De & 15-19th c. & \textit{unk.} & 8.2K & 1K & 1K & 91 \\ 
\bottomrule
\end{tabular}
\label{tab:dataset-comparison}
\end{table}

\paragraph{Performance metrics.}
As the majority of works in the field, we assessed the performance of the models using the common Character Error Rate (CER).\footnote{Results with the Word Error Rate will also be reported in Appendix \ref{sec:B_results}.} This ensures consistency with existing research and facilitates a more direct comparison with our findings.
Additionally, we computed the Expected Calibration Error (ECE) for each model, following the definition in \cite{ayllon_calibration_2024}. We aim to assess not only the accuracy of the models but also their capacity to generate calibrated probability outputs. This is critical for understanding model behavior and uncertainty when applied to unseen OOD data, thereby offering a more comprehensive evaluation of model performance.

\paragraph{Implementation details.}
Each model was trained from scratch for 500 epochs, with the best model saved based on validation performance (CER) for the corresponding domain. Training is always performed with a single source domain, and the run is stopped if the CER on the validation set does not improve at least 0.1 within 100 epochs. We implemented a comprehensive set of typical data augmentation techniques from HTR literature \cite{Luo2020LearnTA, Retsinas_best_practices_2022}: rotation, dilation, erosion, random perspective, elastic transformation, shearing, and Gaussian noise, applied with a probability of 0.5. Lastly, we grayscaled all images as in \cite{multidimensional_recurrent_puig_2017}. Hyperparameters such as architecture, input size, batch size, optimizers, and schedulers were consistent with the original configurations. Further details on data augmentation and hyperparameters are provided in Appendix \ref{sec:A_hyperparameters}. Given that each dataset contains a unique alphabet, we combined all alphabets using Unicode\footnote{\url{https://pypi.org/project/Unidecode/}} to standardize evaluations, resulting in a vocabulary of 94 characters including special symbols for the beginning of sequence [BOS], padding [PAD], unknown [UNK], and the end of sequence [EOS]. No post-processing techniques or lexicon-based predictions were applied to the outputs of the models.

\begin{table*}[ht!]
\caption{In-distribution (ID) and out-of-distribution (OOD) results (CER \%) for HTR models across datasets. The OOD result is reported from the best-performing source. Results marked with $\ast$ indicate outliers, meaning that the model did not converge in the ID setting. Average results (bottom row) are computed filtering out outliers. $\dagger$ denotes architectures implemented from the papers (no code provided).}
\label{tab:results_id_ood}

\renewcommand{\arraystretch}{1.2}   
\setlength{\tabcolsep}{2pt}         
\footnotesize                       

\centering
\resizebox{\textwidth}{!}{%
\begin{tabular}{@{}lcccccccccccccccc@{}}
\toprule
\multirow{2}{*}{\textbf{Dataset}} 
& \multicolumn{2}{c}{\textbf{CRNN} \cite{multidimensional_recurrent_puig_2017}} 
& \multicolumn{2}{c}{\textbf{VAN} \cite{endtoend_coquenet_2022}}
& \multicolumn{2}{c}{\textbf{C-SAN$^\dagger$} \cite{arce_self-attention_2022}}
& \multicolumn{2}{c}{\textbf{HTR-VT} \cite{li_htr-vt_2025}}
& \multicolumn{2}{c}{\textbf{Kang$^\dagger$} \cite{pay_attention_kang_2022}}
& \multicolumn{2}{c}{\textbf{Michael$^\dagger$} \cite{evaluating_michael_2019}}
& \multicolumn{2}{c}{\textbf{LT$^\dagger$} \cite{light_barrere_2022}}
& \multicolumn{2}{c}{\textbf{VLT$^\dagger$} \cite{training_barrere_2024}} \\
\cmidrule(lr){2-3} \cmidrule(lr){4-5} \cmidrule(lr){6-7} \cmidrule(lr){8-9}
\cmidrule(lr){10-11} \cmidrule(lr){12-13} \cmidrule(lr){14-15} \cmidrule(lr){16-17}
& ID & OOD
& ID & OOD
& ID & OOD
& ID & OOD
& ID & OOD
& ID & OOD
& ID & OOD
& ID & OOD \\
\midrule

IAM 
& 6.4 & 34.9$^{\text{\textcolor{red}{(+28.5)}}}$
& 6.6 & 28.6$^{\text{\textcolor{red}{(+22.0)}}}$
& 15.0 & 31.5$^{\text{\textcolor{red}{(+16.6)}}}$
& 5.8 & 33.7$^{\text{\textcolor{red}{(+27.9)}}}$
& 8.0 & 42.1$^{\text{\textcolor{red}{(+34.1)}}}$
& 7.5 & 49.1$^{\text{\textcolor{red}{(+41.6)}}}$
& 7.9 & 42.0$^{\text{\textcolor{red}{(+34.1)}}}$
& 8.9 & 41.3$^{\text{\textcolor{red}{(+32.4)}}}$ \\

Rimes 
& 3.7 & 25.0$^{\text{\textcolor{red}{(+21.2)}}}$
& 5.6 & 21.3$^{\text{\textcolor{red}{(+15.6)}}}$
& 12.0 & 29.8$^{\text{\textcolor{red}{(+19.8)}}}$
& 7.9 & 28.3$^{\text{\textcolor{red}{(+20.4)}}}$
& 5.7 & 32.0$^{\text{\textcolor{red}{(+26.3)}}}$
& 6.9 & 35.5$^{\text{\textcolor{red}{(+28.6)}}}$
& 5.0 & 30.8$^{\text{\textcolor{red}{(+25.8)}}}$
& 5.1 & 29.4$^{\text{\textcolor{red}{(+24.3)}}}$ \\

G.W. 
& 8.2 & 31.1$^{\text{\textcolor{red}{(+22.9)}}}$
& 9.3 & 32.0$^{\text{\textcolor{red}{(+22.7)}}}$
& 9.0 & 49.8$^{\text{\textcolor{red}{(+40.8)}}}$
& 34.9$^{\ast}$ & 38.6$^{\text{\textcolor{red}{}}}$
& 78.4$^{\ast}$ & 44.0$^{\text{\textcolor{red}{}}}$
& 53.8$^{\ast}$ & 43.6$^{\text{\textcolor{red}{}}}$
& 79.6$^{\ast}$ & 32.3$^{\text{\textcolor{red}{}}}$
& 25.2$^{\ast}$ & 32.1$^{\text{\textcolor{red}{(+6.9)}}}$ \\

Bentham 
& 4.7 & 25.3$^{\text{\textcolor{red}{(+20.6)}}}$
& 7.4 & 26.6$^{\text{\textcolor{red}{(+19.2)}}}$
& 10.0 & 38.9$^{\text{\textcolor{red}{(+28.9)}}}$
& 8.4 & 33.3$^{\text{\textcolor{red}{(+24.9)}}}$
& 8.5 & 39.4$^{\text{\textcolor{red}{(+30.9)}}}$
& 8.5 & 43.5$^{\text{\textcolor{red}{(+34.9)}}}$
& 6.0 & 33.8$^{\text{\textcolor{red}{(+27.8)}}}$
& 6.1 & 33.3$^{\text{\textcolor{red}{(+27.2)}}}$ \\

S.G. 
& 7.2 & 33.6$^{\text{\textcolor{red}{(+26.3)}}}$
& 7.8 & 39.8$^{\text{\textcolor{red}{(+32.0)}}}$
& 8.6 & 35.0$^{\text{\textcolor{red}{(+26.4)}}}$
& 17.1 & 36.5$^{\text{\textcolor{red}{(+19.3)}}}$
& 78.7$^{\ast}$ & 51.8$^{\text{\textcolor{red}{}}}$
& 76.9$^{\ast}$ & 55.3$^{\text{\textcolor{red}{}}}$
& 12.5 & 37.8$^{\text{\textcolor{red}{(+25.3)}}}$
& 9.2 & 38.7$^{\text{\textcolor{red}{(+29.4)}}}$ \\

Rodrigo 
& 1.7 & 40.9$^{\text{\textcolor{red}{(+39.3)}}}$
& 2.3 & 38.5$^{\text{\textcolor{red}{(+36.2)}}}$
& 3.8 & 45.2$^{\text{\textcolor{red}{(+41.4)}}}$
& 3.9 & 38.5$^{\text{\textcolor{red}{(+34.6)}}}$
& 2.6 & 60.6$^{\text{\textcolor{red}{(+57.9)}}}$
& 3.8 & 65.3$^{\text{\textcolor{red}{(+61.5)}}}$
& 2.0 & 48.4$^{\text{\textcolor{red}{(+46.4)}}}$
& 2.2 & 47.4$^{\text{\textcolor{red}{(+45.3)}}}$ \\

$\text{ICFHR}_{\text{2016}}$ 
& 5.2 & 78.7$^{\text{\textcolor{red}{(+73.5)}}}$
& 7.5 & 75.3$^{\text{\textcolor{red}{(+67.8)}}}$
& 17.0 & 83.4$^{\text{\textcolor{red}{(+66.4)}}}$
& 11.6 & 79.6$^{\text{\textcolor{red}{(+68.0)}}}$
& 7.8 & 92.6$^{\text{\textcolor{red}{(+84.8)}}}$
& 9.5 & 85.2$^{\text{\textcolor{red}{(+75.7)}}}$
& 5.9 & 90.5$^{\text{\textcolor{red}{(+84.6)}}}$
& 6.0 & 85.1$^{\text{\textcolor{red}{(+79.1)}}}$ \\
\midrule

\textbf{Average}
& \textbf{5.3} & 38.5$^{\text{\textcolor{red}{(+33.2)}}}$
& 6.7 & \textbf{37.4}$^{\text{\textcolor{red}{(+30.8)}}}$
& 9.7 & 44.8$^{\text{\textcolor{red}{(+35.4)}}}$
& 7.5 & 41.2$^{\text{\textcolor{red}{(+33.7)}}}$
& 6.5 & 51.8$^{\text{\textcolor{red}{(+45.3)}}}$
& 7.2 & 53.9$^{\text{\textcolor{red}{(+46.7)}}}$
& 6.5 & 45.1$^{\text{\textcolor{red}{(+38.5)}}}$
& 6.3 & 43.9$^{\text{\textcolor{red}{(+37.6)}}}$ \\
\bottomrule
\end{tabular}%
}
\end{table*}

\section{Practical out-of-distribution insights}
\label{sec:practical_analysis}
We first present the experiments conducted to explore the practical implications of our large-scale study on generalization in HTR. Specifically, we aim to answer the following key questions regarding the studied architectures: (1) Is there a model or alignment strategy that consistently outperforms others in terms of OOD performance? (2) How do model selection and capacity impact OOD performance? Lastly, considering the increasing use of synthetic data in HTR models, we conduct experiments to address the following: (1) Which model or alignment makes the best use of synthetic data to predict real HTR data? and (2) How does the internal Language Model (LM) impact HTR performance?

\subsection{HTR performance}
The results of the first experiments are reported in Table \ref{tab:results_id_ood}, where the rows represent the target domain, the first column indicates the ID error, and the OOD column shows the best error achieved from any other source domain. In this scenario, the typical ID model selection is applied by choosing the one with the best validation error in the source domain. 
The average performance of both models in the two scenarios (ID, OOD) is presented in the last row, with the best results highlighted in bold. We compare three state-of-the-art VLMs \cite{liu2024llavanext, Peng2023Kosmos2GM, instructblip} and TrOCR \cite{trocr_li_2023} in a zero-shot setting in Appendix \ref{sec:B_results}, highlighting their limitations for HTR.



\paragraph{OOD results are terrible.}
OOD results are poor across all architectures and alignment types, with CER values ranging from 37.4\% to 53.9\%—error rates that are notably high for a transcription system. The model that performs best in terms of generalization is VAN, showing only a slight absolute improvement of 1\% over the second-best model (CRNN), which also uses CTC alignment. When comparing the best model (VAN) to the worst model (Michael) in terms of generalization, the absolute improvement is only 16.5\%. The average difference between ID and OOD CERs is 37.6\%, posing that generalization gaps in HTR remain significant even under the best OOD conditions.

\begin{figure}[h!]
\centering
\includegraphics[width=1.0\linewidth]{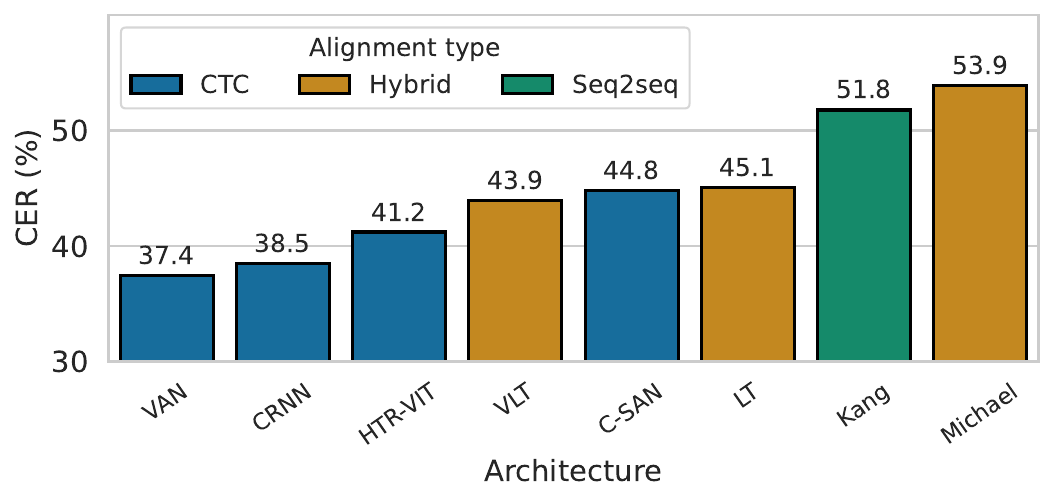}
\caption{Average CER in the out-of-distribution (OOD) scenario where each color represents the alignment type (sorted by performance).}
\label{fig:ranked_CER_alignement}
\end{figure}

\paragraph{Alignment type does not really matter, but choose CTC.}
OOD performance is ordered and color-coded according to the alignment type in Fig.~\ref{fig:ranked_CER_alignement}. While the observed differences are not substantial, a macro-level analysis reveals that models utilizing CTC alignment exhibit the best generalization performance, followed by hybrid approaches, and finally, purely Seq2Seq models. The results suggest that, within the DG scenario, a CTC-alignment model is the most effective choice. Notably, even the model with the poorest OOD performance (C-SAN) is comparable to the second-best alignment type (hybrid approach, VLT), with CER performances of 44.8\% and 43.9\%, respectively. Grouping models by their decoding method—CTC or autoregressive (AR)—we observe that CTC-based models achieve an average CER of 40.4, while AR models have a CER of 48.7, resulting in a relative drop of 20\%.

\subsection{Leveraging synthetic data}
\begin{figure}
    \centering
    \includegraphics[width=1.0\linewidth]{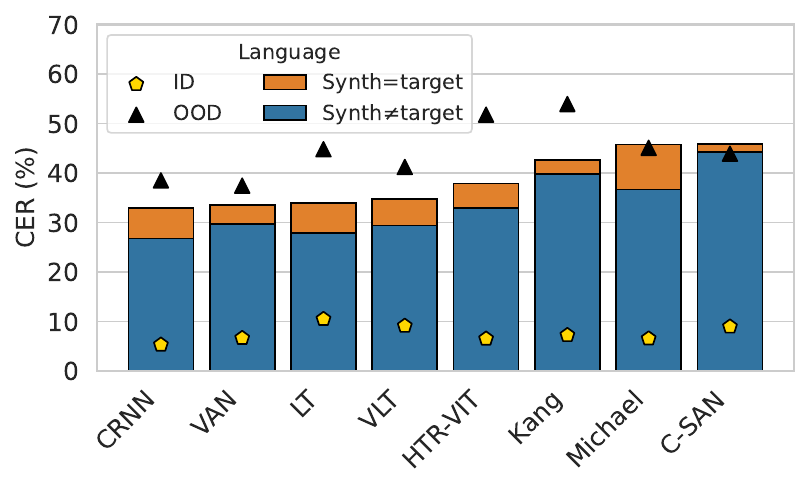}
    \caption{Average performance of the HTR models trained with synthetic data across real historical datasets. Blue bars represent performance when training with the same language as target, whereas orange represents the best result with any other language. ID (green) and OOD (black) performances training with real data are provided as references.}
    \label{fig:rank_synthetic}
\end{figure}

This section investigates how HTR models utilize synthetic data for predicting on the typical HTR datasets. The study has two primary objectives: (1) to assess the effectiveness of each architecture in leveraging synthetic data relative to real data, and (2) to quantify the influence of the underlying language model (LM) on the overall performance. We designed controlled experiments to evaluate the effect of synthetic text generation across multiple languages on real datasets' performances. We generated synthetic 100K text lines randomly selected from WIT dataset \cite{wit_dataset} in English, French, Spanish, German, and Latin using 4,000 publicly available fonts\footnote{Fonts were sourced from open-access repositories like 1001fonts.com.} with handwritten styles ensuring an equal number of lines for each language. We excluded HTG methods \cite{vatr_vanherle_2024, bhunia_handwritting_transformers_2021, handwritten_pippi_2023} as they rely on labeled data targeting ID performance, limiting their applicability to DG. For this scenario, we compare the performance of models trained on synthetic datasets rendered in the same language as the target versus those trained on a different language. To ensure fair comparisons, we consistently selected the best-performing models based on validation results for the target dataset regarding the language used. The results are shown in Fig. \ref{fig:rank_synthetic}, where results are sorted from best to worst (left-to-right) to facilitate visualization. We also report the full results with synthetic data in Appendix \ref{sec:B_results} (Table \ref{tab:all_results_synth}).

\paragraph{Synthetic data (slightly) improves  OOD, even blindly.}
The first observation in Fig. \ref{fig:rank_synthetic} is that using synthetic data consistently improves OOD performance compared to real data, even when a different language is used from the target domain.
Although the CER remains high, ranging from 32.9 to 45.7, models generalize an average of 6 points better when using synthetic data. In the best case, where the target domain language is known, models show an absolute gain of over 11 CER points.

\paragraph{Choose CTC again (AR-models are more biased).} 
Fig. \ref{fig:rank_synthetic} shows that the model that best utilizes synthetic data is the CRNN architecture, followed by VAN. While the differences are less pronounced compared to the OOD scenario using real data, the conclusion remains the same: CTC alignment is still the best option. When grouping models by their decoding method, whether CTC or autoregressive (AR), we find that the relative decrease compared to training with synthetic data---same or a different language---is 25\% for AR and 19\% for CTC. This intuitively reflects the nature and bias of AR models: they are explicitly trained to compute an LM, unlike CTC models, making the latter more robust to language variations.

\subsection{Model selection and capacity} \label{subsec_model-selection}
\begin{figure}
\centering
    \includegraphics[width=1.0\linewidth]{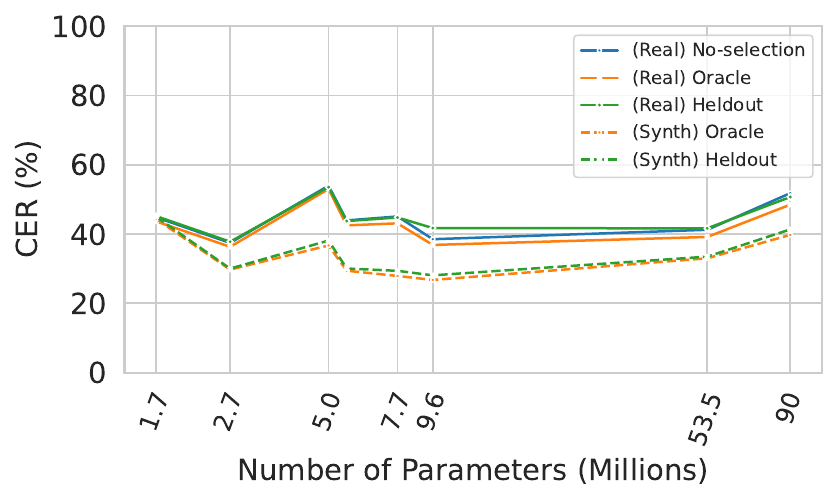}
    \caption{Average CER (\%) in out-of-distribution (OOD) performance for HTR architectures according to the number of parameters (millions). X-axis is in log-scale for better visualization.}
    \label{fig:model_parameters}
\end{figure}

In this section, we evaluate the performance of HTR models through two key factors from the DG literature: model selection \cite{in_search_lost_dg_2021} and model complexity \cite{gouk2024limitationsgeneralpurposedomain}. We compare three OOD model selection strategies inspired by DG studies: (1) \emph{No-selection} (ID selection), where the best model is chosen based on source validation performance (standard i.i.d. setting, as shown in Table \ref{tab:results_id_ood}) used as a baseline; (2) \emph{Heldout selection}, where model selection is based on the average performance across all validation data sources excluding the target dataset \cite{in_search_lost_dg_2021}; (3) \emph{Oracle-based selection}, which sets an upper performance bound by selecting the best model based on validation results from the target dataset \cite{in_search_lost_dg_2021}. As with prior analyses, OOD performance reflects the best result, regardless of the source domain used for training. We present the results in Fig. \ref{fig:model_parameters}, where the capacity of the HTR models (measured in millions of parameters) is compared to their OOD performance across the three model selection methods using real and synthetic data. 

\paragraph{Model selection has no impact.} 
Fig.~\ref{fig:model_parameters} demonstrates that the choice of selection method has a negligible impact on OOD performance in the HTR models, with the results across the various selection strategies being practically identical. Contrary to findings in the broader (DG) literature \cite{in_search_lost_dg_2021}, the oracle-based selection method does not yield substantial advantages over other strategies. These results, when applied to the oracle model, suggest that in the DG scenario, no ``sweet spot'' exists for selecting a model that consistently exhibits superior generalization performance.

\paragraph{More capacity is not useful.} 
It can also be observed from Fig.~\ref{fig:model_parameters} that an increase in model capacity (measured by the number of parameters) does not correspond to an improvement in the generalization capabilities in OOD scenarios, suggesting that increased model complexity does not necessarily translate to better generalization performance.

\section{Factor analysis of out-of-distribution}
In the previous sections, we analyzed the performance of HTR models in OOD scenarios from a practical perspective and compared key aspects of DG literature. In this section, we focus on identifying the hidden factors that most influence OOD generalization in HTR. To achieve this, we conducted a Factor Analysis \cite{general_spearman_1904, confirmatory_wood_2008, frontend_dehak_2011}, where the goal is to find the minimum set of latent variables, or factors, sufficient to explain the data and establish dependencies between features. The features in this analysis consist of the following base metrics that could potentially explain the OOD performance \cite{assaying_ood_2024, accuracy_miller_2021}: model capacity (measured by the number of parameters), ID and OOD errors (measured by CER), and model ID and OOD calibration errors (ECE). However, given that HTR lies in the intersection between image and text processing, we consider two novel features that must be relevant in this context: visual and textual divergence between domains. 

\paragraph{Visual divergence.}   To quantify the visual differences between domains, we approached this problem as an Anomaly Detection (AD) task \cite{visual_anomaly_yang_2022}. Following the reconstruction-based literature \cite{mvtec_bergmann_2019, improving_bergmann_2018, unsupervisedlearningbased_mei_2018}, we measure whether a sample of a target $T$ is out-of-distribution based on the reconstruction error of an Autoencoder (AE) trained on a source $S$ that we denote as $\phi_{\theta_{S}}$. In this work, we trained a simple convolutional AE ($\phi_{\theta_{S}}$) for each source dataset (details and results in Appendix \ref{sec:C_metrics_fa}), and we compute $\Delta(X_S,X_T)$ as the average reconstruction error (measured in MSE) as:

$$\Delta(X_S, X_T) = \frac{1}{m} \sum_{i=1}^{m} \left\| \phi_{\theta_{S}}(X^{(T)}_{i}) - X^{(T)}_{i}\right\|^2 \\$$
on the target dataset's images $X_T$. Therefore, $\Delta(X_S,X_T)$ estimates how far a dataset deviates from the original source domain distribution. Concretely, $\Delta_S(X_S,X_T)$ quantifies visual divergence between train-test splits from same source $S$ and $\Delta_T(X_S,X_T)$ for different $S$ (train) and $T$ (test).

\paragraph{Textual divergence.} To quantify the divergence between two textual distributions ($Y_S, Y_T$), we considered the averaged KL-divergence across varying n-grams \cite{pippi_how_2023, gouve_kl_div_2011}. Specifically, we compute $\Delta(Y_S,Y_T)$  as the average KL-divergence for each n-gram from $n=1 \dots 5$: 
$$\Delta(Y_S,Y_T) = \frac{1}{n}\sum_{i=1}^{n} \sum_{j \in V_n} D_{KL}^{(n)}(P^{(S)}_j \parallel Q^{(T)}_j)$$ 
where $V_n$ is the set of n-grams for the source vocabulary and $D_{KL}^{(n)}(P^{(S)}_j \parallel Q^{(T)}_j)$ represents the KL divergence between the n-gram $j$ in the source and target distributions.  
Consequently, $\Delta(Y_S,Y_T)$ approximates how ``unlikely'' the text in the target domain is relative to the source domain. In our metrics, we denote $\Delta_L(Y_S,Y_T)$ as the divergence between the source text domain $Y_S$ and a synthetic target domain $T$ (the same ones used in the synthetic experiments) where the language $L$ is known. We refer to $\Delta_{GT}(Y_S,Y_T)$ as the actual divergence (see Appendix \ref{sec:C_metrics_fa}) between the source domain $Y_S$ and the ground truth of the target domain $Y_T$. 

\subsection{Main latent factors}
\begin{figure}
    \centering
    \includegraphics[width=1.0\linewidth]{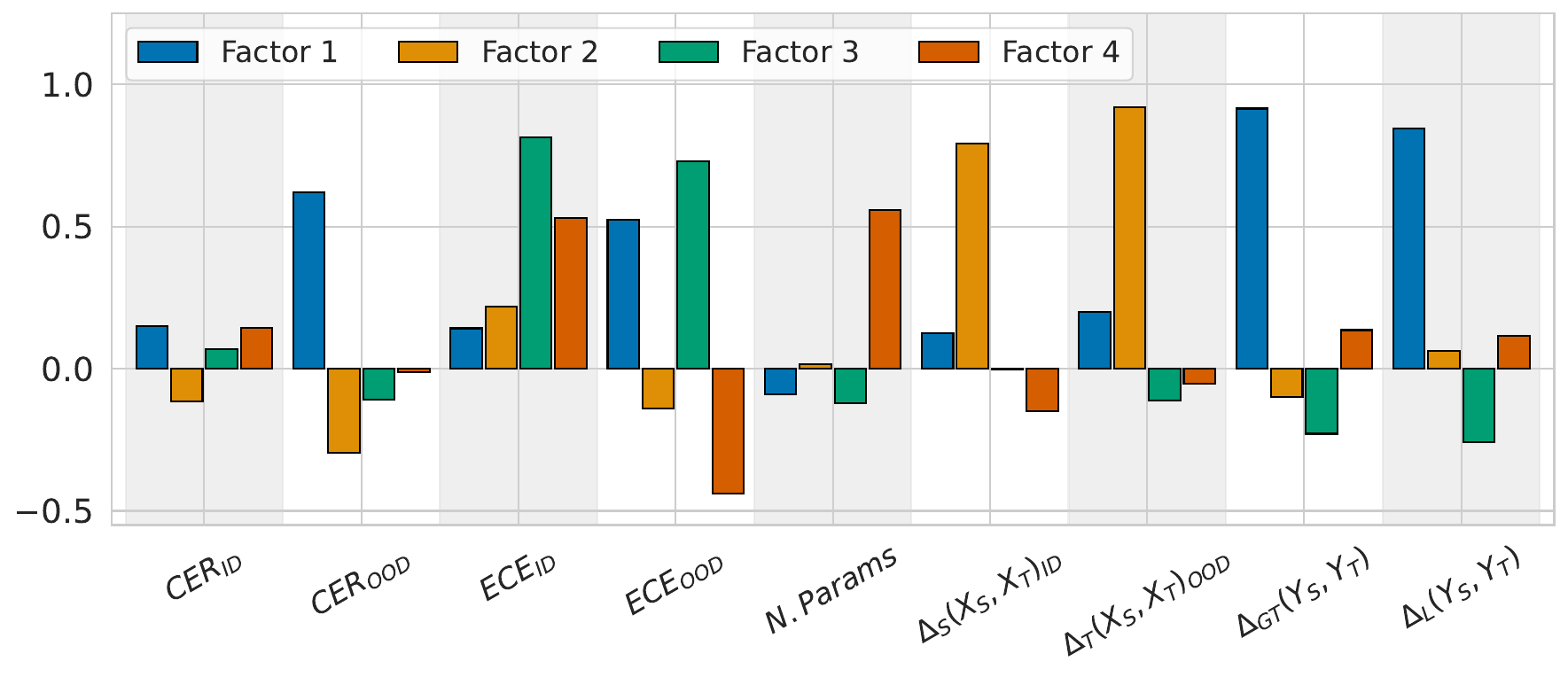}
    \caption{Factor loadings (contributions) of different metrics based on a factor analysis with 4 orthogonal (color-coded) factors: \textcolor{factor1}{Factor 1 (blue)}: Textual divergence; \textcolor{factor2}{Factor 2 (orange)}: Visual divergence; \textcolor{factor3}{Factor 3 (green)}: calibration quality; \textcolor{factor4}{Factor 4 (red):} calibration influenced by model complexity. CER$_\text{OOD}$ is OOD error.}
    \label{fig:factor_analysis}
\end{figure}

Thus, with all metrics collected (model parameters, visual divergence, textual divergence, ID and OOD errors and calibrations) and after applying standard preprocessing to normalize the columns, we extracted the $k$ eigenvectors corresponding to the $k$ eigenvalues $\geq 1$. In our case, $k=4$ (more details in Appendix \ref{sec:D_factor_analysis}). After testing various rotations (although results did not differ significantly) the most interpretable explanation was achieved by \textit{oblimax} rotation. Results are presented in Fig. \ref{fig:factor_analysis}, where color-coded (Pearson's) correlations illustrate how each metric relates to the four hidden factors. 

\paragraph{\textcolor{factor1}{Factor 1}: Textual divergence explains most of the OOD error.}
Factor 1 shows a strong positive correlation with textual metrics for both ground-truth  $\Delta_{GT}(Y_S,Y_T)$ (0.92) and generic texts $\Delta_{L}(Y_S,Y_T)$ (0.84). This suggests that Factor 1 is heavily linked to textual divergences, highlighting it as a significant contributor to OOD error. Noteworthy, Factor 1 shows a moderate positive correlation with OOD error (0.62), implying that greater textual misalignment between domains leads to increased difficulty in generalizing for HTR models. Additionally, Factor 1 shows positive correlations OOD ECE (0.52), suggesting that it may also capture aspects of calibration linked to textual divergence, although to a lesser extent. Overall, these results underscore that alignment in textual content between source and target domain is crucial for reducing OOD error in HTR models.

\paragraph{\textcolor{factor2}{Factor 2}: Visual similarity has a modest influence on OOD Error:} Factor 2 has a high positive correlation with ID reconstruction error $\Delta_S(X_S,X_T)$ (0.8) and OOD reconstruction error $\Delta_T(X_S,X_T)$ (0.92) indicating that this factor predominantly reflects visual divergence between domains. Interestingly, Factor 2 also shows a weak to moderate negative correlation with OOD error (-0.3). This suggests that while visual divergence may not be a dominant factor in predicting OOD error, it does exert a modest influence. This relationship implies that models experience some performance degradation as the visual divergence between source and target domains increases, albeit this effect is weaker compared to the impact of textual divergence. 

\paragraph{\textcolor{factor3}{Factor 3} and {\textcolor{factor4}{Factor 4}}: Calibration quality and model complexity with differing effects on ID and OOD scenarios.}

Factor 3 is strongly correlated with both ID (0.81) and OOD (0.73) ECE, thus representing calibration accuracy. Weak links to model complexity and OOD error imply that this factor mainly affects prediction reliability in OOD scenarios without substantially influencing OOD error itself. Factor 4 correlates positively with ID ECE (0.53) and model parameters (0.56), suggesting it captures ID calibration influenced by model size. It has minimal effect on OOD error, indicating that while increased capacity enhances ID calibration, it does not translate to better generalization.

\subsection{Can we predict OOD performance?}
\begin{figure}
    \centering
    \includegraphics[width=1.0\linewidth]{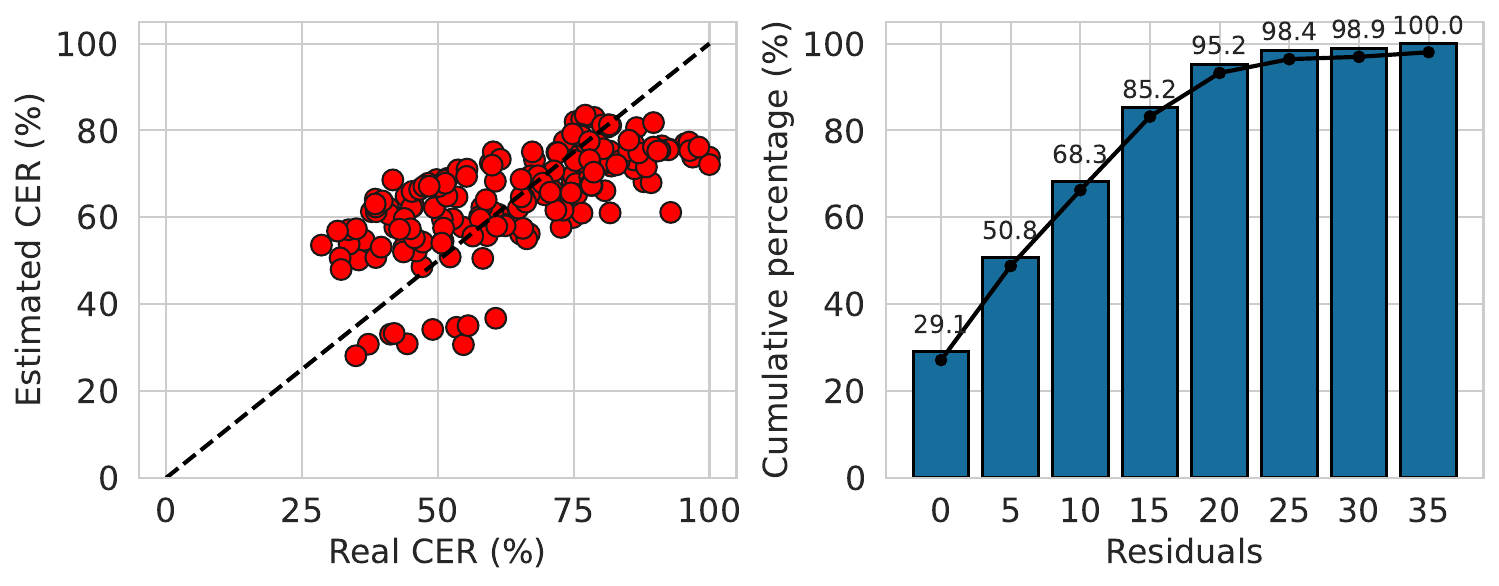}
    \caption{Left: Representation of the estimated vs. real CER values (MSE of 10.9 on average). Right: Cumulative distribution of grouped residuals. Approximately 70\% of predictions yield an error below 10 points of CER.}
    \label{fig:residuals_distr}
\end{figure}

We established that the factors most affecting the generalization of models are those that quantify textual divergences. This section addresses a more practical question: Can OOD error be estimated from the proxy metrics investigated? To analyze this, we calculated the expected error using metrics that do not require target labels: ID error, ID ECE, number of parameters, reconstruction errors (ID and OOD), and KL divergence. The results are displayed in Fig.~\ref{fig:residuals_distr} (left), where the actual CER (X-axis) is plotted against the expected CER (Y-axis). The deviation from the black line represents the absolute error between the predicted and actual OOD values with an average deviation of 10.9 points, measured by Mean Absolute Error (MAE). We observe a clear positive correlation between the real and estimated CER values, suggesting that the model is generally successful in capturing the error trends, with higher predicted CER values tending to correspond with higher real ones. From a practical standpoint, the right plot in Fig.~\ref{fig:residuals_distr} presents the residuals distribution (measured by MAE), highlighting that nearly 30\% of predictions have residuals between 0 and 5 and thus demonstrating high accuracy for a significant portion of samples. By the time the residuals reach the 10–15 range, about 70\% of the predictions have been accounted for, indicating that most predictions are reasonably close to the true CER. 

\section{Conclusions}
\label{sec:conclusions}
This paper provides a comprehensive analysis of generalization in Handwritten Text Recognition (HTR) models, addressing the significant gap in understanding performance in out-of-distribution (OOD) scenarios. We conduct the first large-scale evaluation of HTR model performance, analyzing OOD results across multiple datasets and architectures, first providing practical insights for researchers in the HTR field. These experiments suggested that greater emphasis should be placed on enhancing the generalization capabilities of HTR models, as no architecture or alignment type in the literature currently facilitates effective generalization. Additionally, the results indicated that when utilizing synthetic data, greater benefits are likely to be achieved with architectures based on alignment CTC. Furthermore, leveraging proxy metrics through factor analysis, we identified that the primary factor contributing to OOD error is the textual divergence between source and target, with a weaker contribution of the visual divergence factor. We emphasize that more research studying this last factor has to be done. Finally, we also found that these proxy metrics can robustly predict generalization errors with considerable precision. 

\section*{Acknowledgments}
\label{sec:acknowledgments}
This research was supported by the Spanish Ministry of Science and Innovation through the LEMUR research project (PID2023-148259NB-I00), funded by MCIU/AEI/10.13039/501100011033/FEDER, EU, and the European Social Fund Plus (FSE+). \\
The first author is supported by grant CIACIF/2021/465 from “Programa I+D+i de la Generalitat Valenciana”. 

{
    \small
    \bibliographystyle{ieeenat_fullname}
    \bibliography{main}

\begin{thebibliography}{98}
\providecommand{\natexlab}[1]{#1}
\providecommand{\url}[1]{\texttt{#1}}
\expandafter\ifx\csname urlstyle\endcsname\relax
  \providecommand{\doi}[1]{doi: #1}\else
  \providecommand{\doi}{doi: \begingroup \urlstyle{rm}\Url}\fi

\bibitem[Abdallah et~al.(2020)Abdallah, Hamada, and
  Nurseitov]{attentionbased_abdallah_2020}
Abdelrahman Abdallah, Mohamed~A. Hamada, and Daniyar Nurseitov.
\newblock Attention-based fully gated cnn-bgru for russian handwritten text.
\newblock \emph{Journal of Imaging}, 2020.

\bibitem[Abed et~al.(2010)Abed, Märgner, and
  Blumenstein]{international_abed_2010}
Haikal~El Abed, Volker Märgner, and Michael Blumenstein.
\newblock {International Conference on Frontiers in Handwriting Recognition
  (ICFHR 2010) - Competitions Overview}.
\newblock 2010.

\bibitem[Aberdam et~al.(2021)Aberdam, Litman, Tsiper, Anschel, Slossberg,
  Mazor, Manmatha, and Perona]{sequencetosequence_aberdam_2021}
Aviad Aberdam, Ron Litman, Shahar Tsiper, Oron Anschel, Ron Slossberg, Shai
  Mazor, R. Manmatha, and Pietro Perona.
\newblock Sequence-to-sequence contrastive learning for text recognition.
\newblock \emph{Computer Vision and Pattern Recognition}, 2021.

\bibitem[Aradillas et~al.(2021{\natexlab{a}})Aradillas, Murillo-Fuentes, and
  Olmos]{aradillas_boosting_2021}
Jose~Carlos Aradillas, Juan~Jose Murillo-Fuentes, and Pablo~M. Olmos.
\newblock Boosting {Offline} {Handwritten} {Text} {Recognition} in {Historical}
  {Documents} {With} {Few} {Labeled} {Lines}.
\newblock \emph{IEEE Access}, 9:\penalty0 76674--76688, 2021{\natexlab{a}}.

\bibitem[Aradillas et~al.(2021{\natexlab{b}})Aradillas, Murillo-Fuentes, and
  Olmos]{boosting_aradillas_2021}
Jose~Carlos Aradillas, Juan~Jose Murillo-Fuentes, and Pablo~M. Olmos.
\newblock Boosting offline handwritten text recognition in historical documents
  with few labeled lines.
\newblock \emph{IEEE Access}, 2021{\natexlab{b}}.

\bibitem[Aradillas~Jaramillo et~al.(2018)Aradillas~Jaramillo, Murillo-Fuentes,
  and M.~Olmos]{aradillas_jaramillo_boosting_2018}
Jose~Carlos Aradillas~Jaramillo, Juan~Jose Murillo-Fuentes, and Pablo M.~Olmos.
\newblock Boosting {Handwriting} {Text} {Recognition} in {Small} {Databases}
  with {Transfer} {Learning}.
\newblock In \emph{2018 16th {International} {Conference} on {Frontiers} in
  {Handwriting} {Recognition} ({ICFHR})}, pages 429--434, Niagara Falls, NY,
  USA, 2018. IEEE.

\bibitem[Ayllon et~al.(2024)Ayllon, Castellanos, and
  Calvo-Zaragoza]{ayllon_calibration_2024}
Eric Ayllon, Francisco~J. Castellanos, and Jorge Calvo-Zaragoza.
\newblock Analysis of the calibration of handwriting text recognition
  models.
\newblock In \emph{Document Analysis and Recognition - ICDAR 2024}, pages
  139--155, Cham, 2024. Springer Nature Switzerland.

\bibitem[Baek et~al.(2022)Baek, Jiang, Raghunathan, and
  Kolter]{agreementontheline_baek_2022}
Christina Baek, Yiding Jiang, Aditi Raghunathan, and Zico Kolter.
\newblock Agreement-on-the-line: Predicting the performance of neural networks
  under distribution shift.
\newblock \emph{Neural Information Processing Systems}, 2022.

\bibitem[Bahdanau et~al.(2015)Bahdanau, Cho, and Bengio]{Bahdanau:ICLR:2015}
Dzmitry Bahdanau, Kyunghyun Cho, and Yoshua Bengio.
\newblock Neural machine translation by jointly learning to align and
  translate.
\newblock In \emph{3rd International Conference on Learning Representations,
  {ICLR} 2015, San Diego, CA, USA, May 7-9, 2015, Conference Track
  Proceedings}, 2015.

\bibitem[Barrere et~al.(2022)Barrere, Soullard, Lemaitre, and
  Coüasnon]{light_barrere_2022}
Killian Barrere, Yann Soullard, Aurélie Lemaitre, and Bertrand Coüasnon.
\newblock A light transformer-based architecture for handwritten text
  recognition.
\newblock 2022.

\bibitem[Barrere et~al.(2024)Barrere, Soullard, Lemaitre, and
  Coüasnon]{training_barrere_2024}
Killian Barrere, Yann Soullard, Aurélie Lemaitre, and Bertrand Coüasnon.
\newblock Training transformer architectures on few annotated data: an
  application to historical handwritten text recognition.
\newblock \emph{International Journal on Document Analysis and Recognition
  (IJDAR)}, 2024.

\bibitem[Beery et~al.(2018)Beery, Horn, and Perona]{recognition_beery_2018}
Sara Beery, Grant~Van Horn, and P. Perona.
\newblock Recognition in terra incognita.
\newblock \emph{European Conference on Computer Vision}, 2018.

\bibitem[Ben-David et~al.(2010)Ben-David, Blitzer, Crammer, Kulesza, Pereira,
  and Vaughan]{theory_bendavid_2010}
Shai Ben-David, John Blitzer, Koby Crammer, Alex Kulesza, Fernando Pereira, and
  Jennifer~Wortman Vaughan.
\newblock A theory of learning from different domains.
\newblock \emph{Machine-mediated learning}, 2010.

\bibitem[Bergmann et~al.(2018)Bergmann, Löwe, Fauser, Sattlegger, and
  Steger]{improving_bergmann_2018}
Paul Bergmann, Sindy Löwe, Michael Fauser, David Sattlegger, and C. Steger.
\newblock Improving unsupervised defect segmentation by applying structural
  similarity to autoencoders.
\newblock \emph{VISIGRAPP}, 2018.

\bibitem[Bergmann et~al.(2019)Bergmann, Fauser, Sattlegger, and
  Steger]{mvtec_bergmann_2019}
Paul Bergmann, Michael Fauser, David Sattlegger, and C. Steger.
\newblock Mvtec ad — a comprehensive real-world dataset for unsupervised
  anomaly detection.
\newblock \emph{Computer Vision and Pattern Recognition}, 2019.

\bibitem[Bhunia et~al.(2021{\natexlab{a}})Bhunia, Ghose, Kumar, Chowdhury,
  Sain, and Song]{bhunia_metahtr_2021}
Ayan~Kumar Bhunia, Shuvozit Ghose, Amandeep Kumar, Pinaki~Nath Chowdhury,
  Aneeshan Sain, and Yi-Zhe Song.
\newblock {MetaHTR}: {Towards} {Writer}-{Adaptive} {Handwritten} {Text}
  {Recognition}.
\newblock In \emph{2021 {IEEE}/{CVF} {Conference} on {Computer} {Vision} and
  {Pattern} {Recognition} ({CVPR})}, pages 15825--15834, Nashville, TN, USA,
  2021{\natexlab{a}}. IEEE.

\bibitem[Bhunia et~al.(2021{\natexlab{b}})Bhunia, Khan, Cholakkal, Anwer, Khan,
  and Shah]{bhunia_handwritting_transformers_2021}
Ankan~Kumar Bhunia, Salman Khan, Hisham Cholakkal, Rao~Muhammad Anwer,
  Fahad~Shahbaz Khan, and Mubarak Shah.
\newblock { Handwriting Transformers }.
\newblock In \emph{2021 IEEE/CVF International Conference on Computer Vision
  (ICCV)}, pages 1066--1074, Los Alamitos, CA, USA, 2021{\natexlab{b}}. IEEE
  Computer Society.

\bibitem[Bhunia et~al.(2021{\natexlab{c}})Bhunia, Sain, Chowdhury, and
  Song]{bhunia_text_2021}
Ayan~Kumar Bhunia, Aneeshan Sain, Pinaki~Nath Chowdhury, and Yi-Zhe Song.
\newblock Text is {Text}, {No} {Matter} {What}: {Unifying} {Text} {Recognition}
  using {Knowledge} {Distillation}.
\newblock In \emph{2021 {IEEE}/{CVF} {International} {Conference} on {Computer}
  {Vision} ({ICCV})}, pages 963--972, Montreal, QC, Canada, 2021{\natexlab{c}}.
  IEEE.

\bibitem[Causer and Wallace(2012)]{bentham_causer2012building}
Tim Causer and Valerie Wallace.
\newblock Building a volunteer community: Results and findings from transcribe
  bentham.
\newblock \emph{Digital Humanities Quarterly}, 6\penalty0 (2), 2012.

\bibitem[Coquenet et~al.(2022)Coquenet, Chatelain, and
  Paquet]{endtoend_coquenet_2022}
Denis Coquenet, Clement Chatelain, and Thierry Paquet.
\newblock End-to-end handwritten paragraph text recognition using a vertical
  attention network.
\newblock 2022.

\bibitem[Coquenet et~al.(2023)Coquenet, Chatelain, and
  Paquet]{dan_coquenet_2023}
Denis Coquenet, Clément Chatelain, and Thierry Paquet.
\newblock Dan: a segmentation-free document attention network for handwritten
  document recognition.
\newblock 2023.

\bibitem[Dai et~al.(2023)Dai, Li, Li, Tiong, Zhao, Wang, Li, Fung, and
  Hoi]{instructblip}
Wenliang Dai, Junnan Li, Dongxu Li, Anthony Meng~Huat Tiong, Junqi Zhao,
  Weisheng Wang, Boyang Li, Pascale Fung, and Steven Hoi.
\newblock Instructblip: towards general-purpose vision-language models with
  instruction tuning.
\newblock In \emph{Proceedings of the 37th International Conference on Neural
  Information Processing Systems}, Red Hook, NY, USA, 2023. Curran Associates
  Inc.

\bibitem[Dehak et~al.(2011)Dehak, Kenny, Dehak, Dumouchel, and
  Ouellet]{frontend_dehak_2011}
N. Dehak, P. Kenny, Réda Dehak, P. Dumouchel, and P. Ouellet.
\newblock Front-end factor analysis for speaker verification.
\newblock \emph{IEEE Transactions on Audio, Speech, and Language Processing},
  2011.

\bibitem[Diaz et~al.(2021)Diaz, Ingle, Qin, Bissacco, and
  Fujii]{rethinking_diaz_2021}
Daniel~Hernandez Diaz, Reeve Ingle, Siyang Qin, Alessandro Bissacco, and
  Yasuhisa Fujii.
\newblock Rethinking text line recognition models.
\newblock \emph{arXiv}, 2021.

\bibitem[Dosovitskiy et~al.(2020)Dosovitskiy, Beyer, Kolesnikov, Weissenborn,
  Zhai, Unterthiner, Dehghani, Minderer, Heigold, Gelly, Uszkoreit, and
  Houlsby]{Dosovitskiy2020AnII}
Alexey Dosovitskiy, Lucas Beyer, Alexander Kolesnikov, Dirk Weissenborn,
  Xiaohua Zhai, Thomas Unterthiner, Mostafa Dehghani, Matthias Minderer, Georg
  Heigold, Sylvain Gelly, Jakob Uszkoreit, and Neil Houlsby.
\newblock An image is worth 16x16 words: Transformers for image recognition at
  scale.
\newblock \emph{ArXiv}, abs/2010.11929, 2020.

\bibitem[d’Arce et~al.(2022)d’Arce, Norton, Hannuna, and
  Cristianini]{arce_self-attention_2022}
Rafael d’Arce, Terence Norton, Sion Hannuna, and Nello Cristianini.
\newblock Self-attention networks for non-recurrent handwritten text
  recognition.
\newblock In \emph{Frontiers in Handwriting Recognition}, pages 389--403.
  Springer International Publishing, 2022.
\newblock Series Title: Lecture Notes in Computer Science.

\bibitem[Eyuboglu et~al.(2022)Eyuboglu, Varma, Saab, Delbrouck, Lee-Messer,
  Dunnmon, Zou, and R'e]{domino_eyuboglu_2022}
Sabri Eyuboglu, M. Varma, Khaled~Kamal Saab, Jean-Benoit Delbrouck, Christopher
  Lee-Messer, Jared~A. Dunnmon, James~Y. Zou, and Christopher R'e.
\newblock Domino: Discovering systematic errors with cross-modal embeddings.
\newblock \emph{International Conference on Learning Representations}, 2022.

\bibitem[Fujitake(2023)]{dtrocr_fujitake_2023}
Masato Fujitake.
\newblock Dtrocr: Decoder-only transformer for optical character recognition.
\newblock \emph{arXiv.org}, 2023.

\bibitem[Gouk et~al.(2024)Gouk, Bohdal, Li, and
  Hospedales]{gouk2024limitationsgeneralpurposedomain}
Henry Gouk, Ondrej Bohdal, Da Li, and Timothy Hospedales.
\newblock On the limitations of general purpose domain generalisation methods,
  2024.

\bibitem[Gouv{\^{e}}a and Davel(2011)]{gouve_kl_div_2011}
Evandro Gouv{\^{e}}a and Marelie~H. Davel.
\newblock Kullback-leibler divergence-based {ASR} training data selection.
\newblock In \emph{12th Annual Conference of the International Speech
  Communication Association, {INTERSPEECH} 2011, Florence, Italy, August 27-31,
  2011}, pages 2297--2300. {ISCA}, 2011.

\bibitem[Graves and Schmidhuber(2005)]{Bi-LSTMGRAVES2005602}
Alex Graves and Jürgen Schmidhuber.
\newblock Framewise phoneme classification with bidirectional lstm and other
  neural network architectures.
\newblock \emph{Neural Networks}, 18\penalty0 (5):\penalty0 602--610, 2005.
\newblock IJCNN 2005.

\bibitem[Graves et~al.(2006)Graves, Fernández, Gomez, and
  Schmidhuber]{connectionist_graves_2006}
A. Graves, Santiago Fernández, Faustino~J. Gomez, and J. Schmidhuber.
\newblock Connectionist temporal classification: labelling unsegmented sequence
  data with recurrent neural networks.
\newblock \emph{ICML}, 2006.

\bibitem[Gulrajani and Lopez-Paz(2021)]{in_search_lost_dg_2021}
Ishaan Gulrajani and David Lopez-Paz.
\newblock In search of lost domain generalization.
\newblock \emph{International Conference on Learning Representations}, 2021.

\bibitem[Hochreiter and Schmidhuber(1997)]{Hochreiter-LSTM}
Sepp Hochreiter and J{\"{u}}rgen Schmidhuber.
\newblock Long short-term memory.
\newblock \emph{Neural Comput.}, 9\penalty0 (8):\penalty0 1735--1780, 1997.

\bibitem[Ingle et~al.(2019)Ingle, Fujii, Deselaers, Baccash, and
  Popat]{scalable_htr_ingle_2019}
R.~Reeve Ingle, Yasuhisa Fujii, Thomas Deselaers, Jonathan Baccash, and
  Ashok~C. Popat.
\newblock { A Scalable Handwritten Text Recognition System }.
\newblock In \emph{2019 International Conference on Document Analysis and
  Recognition (ICDAR)}, pages 17--24. IEEE Computer Society, 2019.

\bibitem[Jaderberg et~al.(2014)Jaderberg, Simonyan, Vedaldi, and
  Zisserman]{synthetic_jaderberg_2014}
Max Jaderberg, Karen Simonyan, Andrea Vedaldi, and Andrew Zisserman.
\newblock Synthetic data and artificial neural networks for natural scene text
  recognition.
\newblock \emph{arXiv: Computer Vision and Pattern Recognition}, 2014.

\bibitem[Kang et~al.(2020{\natexlab{a}})Kang, Riba, Rusiñol, Fornés, and
  Villegas]{pay_kang_2020}
Lei Kang, Pau Riba, Marçal Rusiñol, Alicia Fornés, and Mauricio Villegas.
\newblock Pay attention to what you read: Non-recurrent handwritten text-line
  recognition.
\newblock \emph{arXiv (Cornell University)}, 2020{\natexlab{a}}.

\bibitem[Kang et~al.(2020{\natexlab{b}})Kang, Rusinol, Fornes, Riba, and
  Villegas]{kang_unsupervised_2020}
Lei Kang, Marcal Rusinol, Alicia Fornes, Pau Riba, and Mauricio Villegas.
\newblock Unsupervised {Adaptation} for {Synthetic}-to-{Real} {Handwritten}
  {Word} {Recognition}.
\newblock In \emph{2020 {IEEE} {Winter} {Conference} on {Applications} of
  {Computer} {Vision} ({WACV})}, pages 3491--3500, Snowmass Village, CO, USA,
  2020{\natexlab{b}}. IEEE.

\bibitem[Kang et~al.(2022)Kang, Riba, Rusiñol, Fornés, and
  Villegas]{pay_attention_kang_2022}
Lei Kang, Pau Riba, Marçal Rusiñol, Alicia Fornés, and Mauricio Villegas.
\newblock Pay attention to what you read: Non-recurrent handwritten text-line
  recognition.
\newblock \emph{Pattern Recognition}, 129:\penalty0 108766, 2022.

\bibitem[Kass and Vats(2022)]{attentionhtr_kass_2022}
Dmitrijs Kass and Ekta Vats.
\newblock Attentionhtr: Handwritten text recognition based on attention
  encoder-decoder networks.
\newblock 2022.

\bibitem[Kermorvant and Louradour(2010)]{rimes_2010}
Christopher Kermorvant and J{\'{e}}r{\^{o}}me Louradour.
\newblock Handwritten mail classification experiments with the rimes database.
\newblock In \emph{International Conference on Frontiers in Handwriting
  Recognition, {ICFHR} 2010, Kolkata, India, 16-18 November 2010}, pages
  241--246. {IEEE} Computer Society, 2010.

\bibitem[Kim et~al.(2017)Kim, Hori, and Watanabe]{joint_hybrid_speech_kim_2017}
Suyoun Kim, Takaaki Hori, and Shinji Watanabe.
\newblock Joint ctc-attention based end-to-end speech recognition using
  multi-task learning.
\newblock In \emph{2017 IEEE International Conference on Acoustics, Speech and
  Signal Processing (ICASSP)}, pages 4835--4839, 2017.

\bibitem[Kohút et~al.(2023)Kohút, Hradiš, and Kišš]{kohut_towards_2023}
Jan Kohút, Michal Hradiš, and Martin Kišš.
\newblock Towards {Writing} {Style} {Adaptation} in {Handwriting}
  {Recognition}, 2023.
\newblock Version Number: 1.

\bibitem[Krishnan and Jawahar(2016)]{generating_krishnan_2016}
Praveen Krishnan and C.~V. Jawahar.
\newblock Generating synthetic data for text recognition.
\newblock \emph{arXiv: Computer Vision and Pattern Recognition}, 2016.

\bibitem[Kumari et~al.(2022)Kumari, Singh, Rathore, Sharma, Kumari, Singh,
  Rathore, and Sharma]{lexicon_kumari_2022}
Lalita Kumari, Sukhdeep Singh, Vaibhav Varish~Singh Rathore, Anuj Sharma,
  Lalita Kumari, Sukhdeep Singh, Vaibhav Varish~Singh Rathore, and Anuj Sharma.
\newblock Lexicon and attention based handwritten text recognition system.
\newblock 2022.

\bibitem[Li et~al.(2023)Li, Lv, Chen, Cui, Lu, Florencio, Zhang, Li, and
  Wei]{trocr_li_2023}
Minghao Li, Tengchao Lv, Jingye Chen, Lei Cui, Yijuan Lu, Dinei Florencio, Cha
  Zhang, Zhoujun Li, and Furu Wei.
\newblock Trocr: Transformer-based optical character recognition with
  pre-trained models.
\newblock \emph{Proceedings of the ... AAAI Conference on Artificial
  Intelligence}, 2023.

\bibitem[Li et~al.(2024)Li, Chen, Tang, and Shen]{li_htr-vt_2025}
Yuting Li, Dexiong Chen, Tinglong Tang, and Xi Shen.
\newblock {HTR}-{VT}: Handwritten text recognition with vision transformer.
\newblock 158:\penalty0 110967, 2024.

\bibitem[Liu et~al.(2024)Liu, Li, Li, Li, Zhang, Shen, and
  Lee]{liu2024llavanext}
Haotian Liu, Chunyuan Li, Yuheng Li, Bo Li, Yuanhan Zhang, Sheng Shen, and
  Yong~Jae Lee.
\newblock Llava-next: Improved reasoning, ocr, and world knowledge, 2024.

\bibitem[Lu et~al.(2019)Lu, Batra, Parikh, and Lee]{vilbert_lu_2019}
Jiasen Lu, Dhruv Batra, Devi Parikh, and Stefan Lee.
\newblock Vilbert: Pretraining task-agnostic visiolinguistic representations
  for vision-and-language tasks.
\newblock \emph{Neural Information Processing Systems}, 2019.

\bibitem[Lu et~al.(2021)Lu, He, Zhu, Zhang, Song, and Xiang]{simpler_lu_2021}
Zhihe Lu, Sen He, Xiatian Zhu, Li Zhang, Yi-Zhe Song, and Tao Xiang.
\newblock Simpler is better: Few-shot semantic segmentation with classifier
  weight transformer.
\newblock 2021.

\bibitem[Luo et~al.(2020)Luo, Zhu, Jin, and Wang]{Luo2020LearnTA}
Canjie Luo, Yuanzhi Zhu, Lianwen Jin, and Yongpan Wang.
\newblock Learn to augment: Joint data augmentation and network optimization
  for text recognition.
\newblock \emph{2020 IEEE/CVF Conference on Computer Vision and Pattern
  Recognition (CVPR)}, pages 13743--13752, 2020.

\bibitem[Marti and Bunke(2002)]{iamdatabase_marti_2002}
Urs-Viktor Marti and Horst Bunke.
\newblock The iam-database: an english sentence database for offline
  handwriting recognition.
\newblock \emph{International Journal on Document Analysis and Recognition},
  2002.

\bibitem[Mei et~al.(2018)Mei, Yang, and
  Yin]{unsupervisedlearningbased_mei_2018}
Shuang Mei, Hua Yang, and Z. Yin.
\newblock An unsupervised-learning-based approach for automated defect
  inspection on textured surfaces.
\newblock \emph{IEEE Transactions on Instrumentation and Measurement}, 2018.

\bibitem[Michael et~al.(2019)Michael, Labahn, Grüning, and
  Zöllner]{evaluating_michael_2019}
Johannes Michael, R. Labahn, Tobias Grüning, and Jochen Zöllner.
\newblock Evaluating sequence-to-sequence models for handwritten text
  recognition.
\newblock \emph{IEEE International Conference on Document Analysis and
  Recognition}, 2019.

\bibitem[Miller et~al.(2021)Miller, Taori, Raghunathan, Sagawa, Koh, Shankar,
  Liang, Carmon, and Schmidt]{accuracy_miller_2021}
John Miller, Rohan Taori, Aditi Raghunathan, Shiori Sagawa, Pang~Wei Koh,
  Vaishaal Shankar, Percy Liang, Y. Carmon, and Ludwig Schmidt.
\newblock Accuracy on the line: on the strong correlation between
  out-of-distribution and in-distribution generalization.
\newblock \emph{International Conference on Machine Learning}, 2021.

\bibitem[Momeni and BabaAli(2023)]{transformerbased_momeni_2023}
Saleh Momeni and B. BabaAli.
\newblock A transformer-based approach for arabic offline handwritten text
  recognition.
\newblock \emph{arXiv.org}, 2023.

\bibitem[Mostafa et~al.(2021)Mostafa, Mohamed, Ashraf, Elbehery, Jamal,
  Khoriba, and Ghoneim]{ocformer_mostafa_2021}
Aly Mostafa, Omar Mohamed, Ali Ashraf, Ahmed Elbehery, Salma Jamal, Ghada
  Khoriba, and A. Ghoneim.
\newblock Ocformer: A transformer-based model for arabic handwritten text
  recognition.
\newblock \emph{2021 International Mobile, Intelligent, and Ubiquitous
  Computing Conference (MIUCC)}, 2021.

\bibitem[Paul et~al.(2023)Paul, Madan, Mishra, Hegde, Kumar, and
  Aggarwal]{weakly_paul_2023}
S. Paul, Gagan Madan, Akankshya Mishra, N. Hegde, Pradeep Kumar, and Gaurav
  Aggarwal.
\newblock Weakly supervised information extraction from inscrutable handwritten
  document images.
\newblock \emph{arXiv}, 2023.

\bibitem[Peng et~al.(2023)Peng, Wang, Dong, Hao, Huang, Ma, and
  Wei]{Peng2023Kosmos2GM}
Zhiliang Peng, Wenhui Wang, Li Dong, Yaru Hao, Shaohan Huang, Shuming Ma, and
  Furu Wei.
\newblock Kosmos-2: Grounding multimodal large language models to the world.
\newblock \emph{ArXiv}, abs/2306.14824, 2023.

\bibitem[Peñarrubia et~al.(2024)Peñarrubia, Valero-Mas, and
  Calvo-Zaragoza]{selfsupervised_pearrubia_2024}
Carlos Peñarrubia, J.~J. Valero-Mas, and Jorge Calvo-Zaragoza.
\newblock Self-supervised learning for text recognition: A critical survey.
\newblock \emph{arXiv.org}, 2024.

\bibitem[Pippi et~al.(2023{\natexlab{a}})Pippi, Cascianelli, and
  Cucchiara]{handwritten_pippi_2023}
Vittorio Pippi, S. Cascianelli, and R. Cucchiara.
\newblock Handwritten text generation from visual archetypes.
\newblock \emph{arXiv.org}, 2023{\natexlab{a}}.

\bibitem[Pippi et~al.(2023{\natexlab{b}})Pippi, Cascianelli, Kermorvant, and
  Cucchiara]{pippi_how_2023}
Vittorio Pippi, Silvia Cascianelli, Christopher Kermorvant, and Rita Cucchiara.
\newblock How to {Choose} {Pretrained} {Handwriting} {Recognition} {Models} for
  {Single} {Writer} {Fine}-{Tuning}, 2023{\natexlab{b}}.
\newblock Version Number: 1.

\bibitem[Poulos and Valle(2021)]{characterbased_poulos_2021}
Jason Poulos and Rafael Valle.
\newblock Character-based handwritten text transcription with attention
  networks.
\newblock \emph{Neural Computing and Applications}, 2021.

\bibitem[Puigcerver(2017)]{multidimensional_recurrent_puig_2017}
Joan Puigcerver.
\newblock Are multidimensional recurrent layers really necessary for
  handwritten text recognition?
\newblock In \emph{14th {IAPR} International Conference on Document Analysis
  and Recognition, {ICDAR} 2017, Kyoto, Japan, November 9-15, 2017}, pages
  67--72. {IEEE}, 2017.

\bibitem[Qiao et~al.(2020)Qiao, Zhao, and Peng]{learning_qiao_2020}
Fengchun Qiao, Long Zhao, and Xi Peng.
\newblock Learning to learn single domain generalization.
\newblock \emph{Computer Vision and Pattern Recognition}, 2020.

\bibitem[Rath et~al.(2004)Rath, Manmatha, and Lavrenko]{washington_2007}
Toni~M. Rath, R. Manmatha, and Victor Lavrenko.
\newblock A search engine for historical manuscript images.
\newblock In \emph{Proceedings of the 27th Annual International ACM SIGIR
  Conference on Research and Development in Information Retrieval}, page
  369–376, New York, NY, USA, 2004. Association for Computing Machinery.

\bibitem[Read(2007)]{read_htr_tablet_2007}
Janet~C. Read.
\newblock A study of the usability of handwriting recognition for text entry by
  children.
\newblock \emph{Interacting with Computers}, 19\penalty0 (1):\penalty0 57--69,
  2007.
\newblock Moving Face-to-Face communication to Web-based systems.

\bibitem[Retsinas et~al.(2022)Retsinas, Sfikas, Gatos, and
  Nikou]{Retsinas_best_practices_2022}
George Retsinas, Giorgos Sfikas, Basilis Gatos, and Christophoros Nikou.
\newblock Best practices for a handwritten text recognition system.
\newblock In \emph{Document Analysis Systems}, pages 247--259, Cham, 2022.
  Springer International Publishing.

\bibitem[Romero et~al.(2013)Romero, Fornés, Serrano, Sánchez, Toselli,
  Frinken, Vidal, and Lladós]{esposalles_2013}
Verónica Romero, Alicia Fornés, Nicolás Serrano, Joan~Andreu Sánchez,
  Alejandro~H. Toselli, Volkmar Frinken, Enrique Vidal, and Josep Lladós.
\newblock The esposalles database: An ancient marriage license corpus for
  off-line handwriting recognition.
\newblock \emph{Pattern Recognition}, 46\penalty0 (6):\penalty0 1658--1669,
  2013.

\bibitem[Sang and Cuong(2019)]{improving_sang_2019}
D.~V. Sang and Le~Tran~Bao Cuong.
\newblock Improving crnn with efficientnet-like feature extractor and
  multi-head attention for text recognition.
\newblock \emph{SoICT 2019}, 2019.

\bibitem[Scherrer(1875)]{saint_gall_scherrer1875verzeichniss}
Gustav Scherrer.
\newblock \emph{Verzeichniss der Handschriften der Stiftsbibliothek von St.
  Gallen}.
\newblock Halle, 1875.

\bibitem[Serrano et~al.(2010)Serrano, Castro, and
  Juan]{serrano-etal-2010-rodrigo}
Nicolas Serrano, Francisco Castro, and Alfons Juan.
\newblock The {RODRIGO} database.
\newblock In \emph{Proceedings of the Seventh International Conference on
  Language Resources and Evaluation ({LREC}'10)}, Valletta, Malta, 2010.
  European Language Resources Association (ELRA).

\bibitem[Soullard et~al.(2019)Soullard, Swaileh, Tranouez, Paquet, and
  Chatelain]{soullard_improving_2019}
Yann Soullard, Wassim Swaileh, Pierrick Tranouez, Thierry Paquet, and Clement
  Chatelain.
\newblock Improving {Text} {Recognition} using {Optical} and {Language} {Model}
  {Writer} {Adaptation}.
\newblock In \emph{2019 {International} {Conference} on {Document} {Analysis}
  and {Recognition} ({ICDAR})}, pages 1175--1180, Sydney, Australia, 2019.
  IEEE.

\bibitem[Spearman(1904)]{general_spearman_1904}
C. Spearman.
\newblock General intelligence objectively determined and measured.
\newblock 1904.

\bibitem[Srinivasan et~al.(2021)Srinivasan, Raman, Chen, Bendersky, and
  Najork]{wit_dataset}
Krishna Srinivasan, Karthik Raman, Jiecao Chen, Michael Bendersky, and Marc
  Najork.
\newblock Wit: Wikipedia-based image text dataset for multimodal multilingual
  machine learning.
\newblock In \emph{Proceedings of the 44th International ACM SIGIR Conference
  on Research and Development in Information Retrieval}, page 2443–2449, New
  York, NY, USA, 2021. Association for Computing Machinery.

\bibitem[Sánchez et~al.(2014)Sánchez, Romero, Toselli, and
  Vidal]{icfhr2014_snchez_2014}
Joan~Andreu Sánchez, Verónica Romero, A. Toselli, and E. Vidal.
\newblock Icfhr2014 competition on handwritten text recognition on
  transcriptorium datasets (htrts).
\newblock \emph{2014 14th International Conference on Frontiers in Handwriting
  Recognition}, 2014.

\bibitem[Sánchez et~al.(2016)Sánchez, Romero, Toselli, and
  Vidal]{icfhr_2016_sanchez}
Joan~Andreu Sánchez, Verónica Romero, Alejandro~H. Toselli, and Enrique
  Vidal.
\newblock Icfhr2016 competition on handwritten text recognition on the read
  dataset.
\newblock In \emph{2016 15th International Conference on Frontiers in
  Handwriting Recognition (ICFHR)}, pages 630--635, 2016.

\bibitem[Sánchez et~al.(2017)Sánchez, Romero, Toselli, Villegas, and
  Vidal]{icdar2017_snchez_2017}
Joan-Andreu Sánchez, Verónica Romero, A. Toselli, M. Villegas, and E. Vidal.
\newblock Icdar2017 competition on handwritten text recognition on the read
  dataset.
\newblock \emph{2017 14th IAPR International Conference on Document Analysis
  and Recognition (ICDAR)}, 2017.

\bibitem[Tan et~al.(2018)Tan, Sun, Kong, Zhang, Yang, and Liu]{survey_tan_2018}
Chuanqi Tan, F. Sun, Tao Kong, Wenchang Zhang, Chao Yang, and Chunfang Liu.
\newblock A survey on deep transfer learning.
\newblock \emph{International Conference on Artificial Neural Networks}, 2018.

\bibitem[Tula et~al.(2023)Tula, Paul, Madan, Garst, Ingle, and
  Aggarwal]{tula_is_2023}
Debapriya Tula, Sujoy Paul, Gagan Madan, Peter Garst, Reeve Ingle, and Gaurav
  Aggarwal.
\newblock Is it an i or an l: {Test}-time {Adaptation} of {Text} {Line}
  {Recognition} {Models}, 2023.
\newblock Version Number: 1.

\bibitem[van~der Werff et~al.(2023)van~der Werff, Dhali, and
  Schomaker]{van_der_werff_writer_2023}
Tobias van~der Werff, Maruf~A. Dhali, and Lambert Schomaker.
\newblock Writer adaptation for offline text recognition: {An} exploration of
  neural network-based methods, 2023.
\newblock Version Number: 1.

\bibitem[Vanherle et~al.(2024)Vanherle, Pippi, Cascianelli, Michiels, Reeth,
  and Cucchiara]{vatr_vanherle_2024}
Bram Vanherle, Vittorio Pippi, S. Cascianelli, Nick Michiels, F. Reeth, and R.
  Cucchiara.
\newblock Vatr++: Choose your words wisely for handwritten text generation.
\newblock \emph{arXiv.org}, 2024.

\bibitem[Vaswani et~al.(2017)Vaswani, Shazeer, Parmar, Uszkoreit, Jones, Gomez,
  Kaiser, and Polosukhin]{TransformerVaswani}
Ashish Vaswani, Noam Shazeer, Niki Parmar, Jakob Uszkoreit, Llion Jones,
  Aidan~N Gomez, \L~ukasz Kaiser, and Illia Polosukhin.
\newblock Attention is all you need.
\newblock In \emph{Advances in Neural Information Processing Systems}. Curran
  Associates, Inc., 2017.

\bibitem[Wang et~al.(2021{\natexlab{a}})Wang, Lan, Liu, Ouyang, and
  Qin]{generalizing_wang_2021}
Jindong Wang, Cuiling Lan, Chang Liu, Yidong Ouyang, and Tao Qin.
\newblock Generalizing to unseen domains: A survey on domain generalization.
\newblock \emph{IEEE Transactions on Knowledge and Data Engineering},
  2021{\natexlab{a}}.

\bibitem[Wang and Deng(2018)]{deep_wang_2018}
Mei~Legam Wang and Weihong Deng.
\newblock Deep visual domain adaptation: A survey.
\newblock \emph{Neurocomputing}, 2018.

\bibitem[Wang et~al.(2021{\natexlab{b}})Wang, Luo, Qiu, Huang, and
  Baktashmotlagh]{learning_wang_2021}
Zijian Wang, Yadan Luo, Ruihong Qiu, Zi Huang, and Mahsa Baktashmotlagh.
\newblock Learning to diversify for single domain generalization.
\newblock \emph{2021 IEEE/CVF International Conference on Computer Vision
  (ICCV)}, 2021{\natexlab{b}}.

\bibitem[Weiss et~al.(2016)Weiss, Khoshgoftaar, and Wang]{survey_weiss_2016}
Karl~R. Weiss, Taghi~M. Khoshgoftaar, and Dingding Wang.
\newblock A survey of transfer learning.
\newblock \emph{Journal of Big Data}, 2016.

\bibitem[Wenzel et~al.(2024)Wenzel, Simon-Gabriel, Kernert, Schiele, Dittadi,
  Horn, Russell, Sch\"{o}lkopf, Gehler, Zietlow, Brox, and
  Locatello]{assaying_ood_2024}
Florian Wenzel, Carl-Johann Simon-Gabriel, David Kernert, Bernt Schiele, Andrea
  Dittadi, Max Horn, Chris Russell, Bernhard Sch\"{o}lkopf, Peter Gehler,
  Dominik Zietlow, Thomas Brox, and Francesco Locatello.
\newblock Assaying out-of-distribution generalization in transfer learning.
\newblock In \emph{Proceedings of the 36th International Conference on Neural
  Information Processing Systems}. Curran Associates Inc., 2024.

\bibitem[Wick et~al.(2021{\natexlab{a}})Wick, Zöllner, and
  Grüning]{rescoring_wick_2021}
Christoph Wick, Jochen Zöllner, and Tobias Grüning.
\newblock Rescoring sequence-to-sequence models for text line recognition with
  ctc-prefixes.
\newblock \emph{arXiv: Computer Vision and Pattern Recognition},
  2021{\natexlab{a}}.

\bibitem[Wick et~al.(2021{\natexlab{b}})Wick, Zöllner, and
  Grüning]{transformer_wick_2021}
C. Wick, Jochen Zöllner, and Tobias Grüning.
\newblock Transformer for handwritten text recognition using bidirectional
  post-decoding.
\newblock \emph{ICDAR}, 2021{\natexlab{b}}.

\bibitem[Wilson and Cook(2018)]{survey_wilson_2018}
Garrett Wilson and D. Cook.
\newblock A survey of unsupervised deep domain adaptation.
\newblock \emph{ACM Transactions on Intelligent Systems and Technology}, 2018.

\bibitem[Wood(2008)]{confirmatory_wood_2008}
Phil Wood.
\newblock Confirmatory factor analysis for applied research.
\newblock 2008.

\bibitem[Yang et~al.(2022{\natexlab{a}})Yang, Xu, Qi, and
  Shi]{visual_anomaly_yang_2022}
Jie Yang, Ruijie Xu, Zhiquan Qi, and Yong Shi.
\newblock Visual anomaly detection for images: A systematic survey.
\newblock \emph{Procedia Computer Science}, 199:\penalty0 471--478,
  2022{\natexlab{a}}.
\newblock The 8th International Conference on Information Technology and
  Quantitative Management (ITQM 2020 \& 2021): Developing Global Digital
  Economy after COVID-19.

\bibitem[Yang et~al.(2022{\natexlab{b}})Yang, Liao, Lu, Wang, Zhu, Luo, Tian,
  and Bai]{reading_yang_2022}
Mingkun Yang, Minghui Liao, Pu Lu, Jing Wang, Shenggao Zhu, Hualin Luo,
  Qingzhen Tian, and X. Bai.
\newblock Reading and writing: Discriminative and generative modeling for
  self-supervised text recognition.
\newblock \emph{ACM Multimedia}, 2022{\natexlab{b}}.

\bibitem[Yang et~al.(2019)Yang, Dai, Yang, Carbonell, Salakhutdinov, and
  Le]{xlnet_yang_2019}
Zhilin Yang, Zihang Dai, Yiming Yang, Jaime~G. Carbonell, Ruslan Salakhutdinov,
  and Quoc~V. Le.
\newblock Xlnet: Generalized autoregressive pretraining for language
  understanding.
\newblock \emph{arXiv: Computation and Language}, 2019.

\bibitem[Zhang et~al.(2019)Zhang, Nie, Liu, Xu, Zhang, and
  Shen]{zhang_sequence--sequence_2019}
Yaping Zhang, Shuai Nie, Wenju Liu, Xing Xu, Dongxiang Zhang, and Heng~Tao
  Shen.
\newblock Sequence-{To}-{Sequence} {Domain} {Adaptation} {Network} for {Robust}
  {Text} {Image} {Recognition}.
\newblock In \emph{2019 {IEEE}/{CVF} {Conference} on {Computer} {Vision} and
  {Pattern} {Recognition} ({CVPR})}, pages 2735--2744, Long Beach, CA, USA,
  2019. IEEE.

\bibitem[Zhou et~al.(2021)Zhou, Liu, Qiao, Xiang, and Loy]{domain_zhou_2021}
Kaiyang Zhou, Ziwei Liu, Y. Qiao, T. Xiang, and Chen~Change Loy.
\newblock Domain generalization: A survey.
\newblock \emph{IEEE Transactions on Pattern Analysis and Machine
  Intelligence}, 2021.

\bibitem[Zhuang et~al.(2019)Zhuang, Qi, Duan, Xi, Zhu, Zhu, Xiong, and
  He]{comprehensive_zhuang_2019}
Fuzhen Zhuang, Zhiyuan Qi, Keyu Duan, Dongbo Xi, Yongchun Zhu, Hengshu Zhu, Hui
  Xiong, and Qing He.
\newblock A comprehensive survey on transfer learning.
\newblock \emph{Proceedings of the IEEE}, 2019.

\end{thebibliography}
}

\clearpage
\setcounter{page}{1}
\maketitlesupplementary

\section{Complementary results}
\label{sec:B_results}
In this section, we provide additional metrics to the Character Error Rate (CER) presented in the main paper. We provide the following information regarding the results: the main table of the paper (Table \ref{tab:results_id_ood}) is presented in terms of Word Error Rate (WER) in Table \ref{tab:results_id_ood_wer} and the dataset source for the best test result obtained for each architecture is reported in Table \ref{tab:source_datasets_tests}. Additionally, Tables 
\ref{tab:all_results_real} and \ref{tab:all_results_synth} present all cross-domain results (both real and synthetic, respectively) for each architecture individually. In Table \ref{tab:all_results_real}, the in-domain (ID) results are displayed on the main diagonals and highlighted in gray. 

We observe in Table \ref{tab:results_id_ood_wer} that the ID results remain generally consistent, comparable to those reported in the literature. However, the performance in the OOD scenario remains significantly poor. Additionally, the numerical values highlight that the WER, being a stricter metric than the CER, amplifies the performance gap in the OOD scenario. While the average ID to OOD gap was 37.6\% in terms of CER, it increases to 60.3\% when measured using WER. Table \ref{tab:source_datasets_tests} presents the best-performing source domain for each target domain in the OOD scenario. The dominance of the IAM dataset is noticeable, accounting for nearly 60\% of cases, followed by Bentham with slightly over 20\%. However, as reported in Section \ref{subsec_model-selection}, the choice of source domain has minimal impact on OOD performance, as all results remain poor even when the target domain is known (oracle scenario). Additionally,  

 Lastly, we compare three state-of-the-art VLMs \cite{liu2024llavanext, Peng2023Kosmos2GM, instructblip} including TrOCR \cite{trocr_li_2023} against the best-reported OOD results in the paper (HTR$_\text{OOD}$ column) in Table \ref{tab:vlm_comparison}. As observed, the zero-shot performance of these models is very low, as they are not originally designed for HTR tasks, with TrOCR being on par with the HTR models on the English datasets. Moreover, there is no established pipeline for effectively applying VLMs to such specific HTR datasets, highlighting the need for further investigation.

\begin{table*}[ht!]
\caption{In-distribution (ID) and out-of-distribution (OOD) results (WER \%) for HTR models across datasets. The OOD result is reported from the best-performing source. Results marked with $\ast$ indicate outliers, meaning that the model did not converge in the ID setting. Average results (bottom row) are computed filtering out outliers. $\dagger$ denotes architectures implemented from the papers (no code provided).}
\label{tab:results_id_ood_wer}
\renewcommand{\arraystretch}{1.4}   
\setlength{\tabcolsep}{2pt}         
\footnotesize
\centering
\resizebox{\textwidth}{!}{%
\begin{tabular}{@{}lcccccccccccccccc@{}}
\toprule
\multirow{2}{*}{\textbf{Dataset}} 
& \multicolumn{2}{c}{\textbf{CRNN} \cite{multidimensional_recurrent_puig_2017}} 
& \multicolumn{2}{c}{\textbf{VAN} \cite{endtoend_coquenet_2022}}
& \multicolumn{2}{c}{\textbf{C-SAN$^\dagger$} \cite{arce_self-attention_2022}}
& \multicolumn{2}{c}{\textbf{HTR-VT} \cite{li_htr-vt_2025}}
& \multicolumn{2}{c}{\textbf{Kang$^\dagger$} \cite{pay_attention_kang_2022}}
& \multicolumn{2}{c}{\textbf{Michael$^\dagger$} \cite{evaluating_michael_2019}}
& \multicolumn{2}{c}{\textbf{LT$^\dagger$} \cite{light_barrere_2022}}
& \multicolumn{2}{c}{\textbf{VLT$^\dagger$} \cite{training_barrere_2024}} \\
\cmidrule(lr){2-3} \cmidrule(lr){4-5} \cmidrule(lr){6-7} \cmidrule(lr){8-9} \cmidrule(lr){10-11} \cmidrule(lr){12-13} \cmidrule(lr){14-15} \cmidrule(lr){16-17}
& ID & OOD
& ID & OOD
& ID & OOD
& ID & OOD
& ID & OOD
& ID & OOD
& ID & OOD
& ID & OOD \\
\midrule
IAM 
& 22.4 & 68.2
& 24.2 & 76.8
& 50.7 & 83.6
& 18.2 & 83.7
& 23.2 & 87.1
& 20.2 & 82.9
& 23.4 & 72.0
& 27.0 & 70.3 \\
Rimes 
& 11.5 & 74.2
& 18.2 & 66.9
& 45.1 & 80.7
& 24.5 & 78.7
& 14.8 & 78.3
& 20.0 & 84.5
& 13.2 & 77.4
& 13.3 & 76.1 \\
G.W. 
& 26.0 & 68.8
& 31.2 & 74.7
& 33.8 & 93.9
& 71.7$^{\ast}$ & 83.3
& 104.3$^{\ast}$ & 77.6
& 80.1$^{\ast}$ & 76.1
& 195.4$^{\ast}$ & 65.7
& 51.4$^{\ast}$ & 65.8 \\
Bentham 
& 13.2 & 54.9
& 20.5 & 62.1
& 30.2 & 82.2
& 24.6 & 70.8
& 19.8 & 68.2
& 19.8 & 69.6
& 14.3 & 60.6
& 15.5 & 60.5 \\
S.G. 
& 31.1 & 96.8
& 33.1 & 95.3
& 41.0 & 96.4
& 59.5 & 97.4
& 203.3$^{\ast}$ & 98.7
& 134.7$^{\ast}$ & 98.8
& 45.0 & 97.9
& 37.6 & 97.3 \\
Rodrigo 
& 8.3 & 92.1
& 12.4 & 92.9
& 22.3 & 97.2
& 16.3 & 92.2
& 11.1 & 96.8
& 15.4 & 96.2
& 9.6 & 94.0
& 10.7 & 94.3 \\
$\text{ICFHR}_{\text{2016}}$ 
& 22.1 & 104.5
& 33.9 & 100.4
& 64.2 & 100.7
& 43.3 & 103.1
& 31.7 & 106.8
& 33.9 & 104.8
& 24.2 & 110.3
& 24.6 & 109.7 \\
\midrule
\textbf{Average} 
& \textbf{19.2} & \textbf{79.9}
& 24.8 & 81.3
& 37.2 & 90.7
& 25.4 & 87.0
& 20.1 & 87.6
& 22.3 & 87.6
& 21.6 & 82.6
& 25.7 & 82.0 \\
\bottomrule
\end{tabular}%
}
\end{table*}

\begin{table*}[ht!]
\centering
\caption{
Best source domain for each target (rows) across all architectures studied in the paper. 
}
\label{tab:source_datasets_tests}
\renewcommand{\arraystretch}{1.4} 
\small 
\setlength{\tabcolsep}{5pt} 
\begin{tabular}{lcccccccc}
\hline

\textbf{Dataset}     & \textbf{CRNN} \cite{multidimensional_recurrent_puig_2017} & \textbf{VAN} \cite{endtoend_coquenet_2022} & \textbf{C-SAN$^\dagger$} \cite{arce_self-attention_2022} & \textbf{HTR-VT} \cite{li_htr-vt_2025} & \textbf{Kang$^\dagger$} \cite{pay_attention_kang_2022}  & \textbf{Michael$^\dagger$} \cite{evaluating_michael_2019} & \textbf{LT$^\dagger$} \cite{light_barrere_2022} & \textbf{VLT$^\dagger$} \cite{training_barrere_2024} \\ \hline

\textbf{IAM}         & Bentham        & Rimes          & Rimes           & Rimes            & Rimes          & Bentham          & Bentham      & Bentham      \\
\textbf{G.W.}        & IAM            & Bentham        & IAM             & Bentham          & IAM            & IAM              & IAM          & IAM          \\
\textbf{Bentham}     & IAM            & IAM            & IAM             & IAM              & IAM            & IAM              & IAM          & IAM          \\
\textbf{Rimes}       & IAM            & IAM            & IAM             & IAM              & IAM            & IAM              & IAM          & IAM          \\
\textbf{S.G.}        & Rodrigo        & IAM            & Rodrigo         & Rodrigo          & IAM            & IAM              & Rodrigo      & Rodrigo      \\
\textbf{Rodrigo}     & IAM            & IAM            & Bentham         & IAM              & IAM            & IAM              & IAM          & IAM          \\
\textbf{ICFHR 2016}  & Bentham        & Bentham        & Bentham         & Bentham          & Rimes          & IAM              & Rimes        & Bentham      \\ \hline
\end{tabular}
\end{table*}

\begin{table}[ht!]
    \caption{Complete CER results in all datasets using real data.}
    \label{tab:all_results_real}
    \renewcommand{\arraystretch}{0.9} 
    \centering
    \tiny
    \setlength{\tabcolsep}{1.2pt} 
    \begin{tabular}{llccccccl}
        \toprule
        Method & S/T & IAM & Rimes & G.W. & Bentham & S.G. & Rodrigo & ICFHR$_{2016}$ \\ \midrule
        \multirow{7}{*}{CRNN \cite{multidimensional_recurrent_puig_2017}} 
        & IAM           & \colorbox{gray!30}{6.4} & 25.0 & 31.1 & 25.3 & 45.5 & 40.9 & 86.2 \\
        & Rimes         & 35.4 & \colorbox{gray!30}{3.7} & 49.0 & 50.2 & 52.3 & 47.1 & 87.9 \\
        & G.W.          & 55.6 & 61.5 & \colorbox{gray!30}{8.2} & 59.2 & 69.3 & 66.2 & 100.0 \\
        & Bentham       & 34.9 & 45.3 & 32.2 & \colorbox{gray!30}{4.7} & 57.8 & 43.8 & 78.7 \\
        & S.G.          & 77.7 & 74.5 & 89.3 & 78.0 & \colorbox{gray!30}{7.2} & 52.8 & 100.0 \\
        & Rodrigo       & 65.7 & 61.4 & 71.8 & 66.3 & 33.6 & \colorbox{gray!30}{1.7} & 85.3 \\
        & ICFHR\(_{2016}\)  & 74.9 & 78.4 & 81.6 & 75.4 & 77.9 & 75.6 & \colorbox{gray!30}{5.2} \\ \midrule

        \multirow{7}{*}{VAN \cite{endtoend_coquenet_2022}}
        & IAM           & \colorbox{gray!30}{6.6} & 21.3 & 34.5 & 26.6 & 39.8 & 38.5 & 82.9 \\
        & Rimes         & 28.6 & \colorbox{gray!30}{5.6} & 46.1 & 45.0 & 47.2 & 43.7 & 88.4 \\
        & G.W.          & 73.7 & 67.4 & \colorbox{gray!30}{9.3} & 59.3 & 67.1 & 69.6 & 100.0 \\
        & Bentham       & 37.2 & 41.7 & 32.0 & \colorbox{gray!30}{7.4} & 49.4 & 38.6 & 75.3 \\
        & S.G.          & 96.1 & 85.0 & 93.1 & 83.1 & \colorbox{gray!30}{7.8} & 57.7 & 100.0 \\
        & Rodrigo       & 76.5 & 70.7 & 78.2 & 68.1 & 41.1 & \colorbox{gray!30}{2.3} & 87.3 \\
        & ICFHR\(_{2016}\)  & 70.8 & 74.8 & 76.4 & 67.8 & 72.1 & 71.4 & \colorbox{gray!30}{7.5} \\ \midrule

        \multirow{7}{*}{C-SAN \cite{arce_self-attention_2022}}
        & IAM           & \colorbox{gray!30}{28.6} & 29.8 & 49.8 & 38.9 & 50.2 & 46.6 & 90.9 \\
        & Rimes         & 31.5 & \colorbox{gray!30}{21.3} & 50.7 & 45.9 & 51.1 & 45.7 & 87.0 \\
        & G.W.          & 60.7 & 60.2 & \colorbox{gray!30}{32.0} & 63.2 & 68.4 & 66.8 & 96.9 \\
        & Bentham       & 54.7 & 51.3 & 58.3 & \colorbox{gray!30}{26.6} & 64.8 & 45.2 & 83.4 \\
        & S.G.          & 75.3 & 72.5 & 81.8 & 75.5 & \colorbox{gray!30}{39.8} & 50.8 & 90.2 \\
        & Rodrigo       & 72.7 & 69.1 & 78.1 & 68.2 & 35.0 & \colorbox{gray!30}{38.5} & 86.9 \\
        & ICFHR\(_{2016}\)  & 78.4 & 81.8 & 86.6 & 78.7 & 86.1 & 81.2 & \colorbox{gray!30}{75.3} \\ \midrule

        \multirow{7}{*}{HTR-VT \cite{li_htr-vt_2025}}
        & IAM           & \colorbox{gray!30}{5.8} & 28.3 & 40.0 & 33.3 & 44.2 & 38.5 & 86.1 \\
        & Rimes         & 33.7 & \colorbox{gray!30}{7.9} & 46.2 & 48.2 & 51.0 & 46.1 & 81.9 \\
        & G.W.          & 70.1 & 73.3 & \colorbox{gray!30}{34.9} & 76.3 & 79.2 & 76.9 & 85.9 \\
        & Bentham       & 44.4 & 49.8 & 38.6 & \colorbox{gray!30}{8.4} & 58.1 & 45.4 & 79.6 \\
        & S.G.          & 78.7 & 78.2 & 89.8 & 78.1 & \colorbox{gray!30}{17.1} & 60.1 & 100.0 \\
        & Rodrigo       & 66.4 & 64.1 & 68.7 & 67.8 & 36.5 & \colorbox{gray!30}{3.9} & 85.4 \\
        & ICFHR\(_{2016}\)  & 74.9 & 77.1 & 78.8 & 73.3 & 76.2 & 71.9 & \colorbox{gray!30}{11.6} \\ \midrule

        \multirow{7}{*}{Kang \cite{pay_attention_kang_2022}}
        & IAM           & \colorbox{gray!30}{8.0} & 32.0 & 44.0 & 39.4 & 51.8 & 60.6 & 96.2 \\
        & Rimes         & 42.1 & \colorbox{gray!30}{5.7} & 65.5 & 63.5 & 62.1 & 65.2 & 92.6 \\
        & G.W.          & 82.8 & 81.8 & \colorbox{gray!30}{78.4} & 77.7 & 78.3 & 77.4 & 100.0 \\
        & Bentham       & 53.4 & 59.7 & 46.6 & \colorbox{gray!30}{8.5} & 80.8 & 71.1 & 95.5 \\
        & S.G.          & 100.0 & 100.0 & 100.0 & 100.0 & \colorbox{gray!30}{78.7} & 86.5 & 100.0 \\
        & Rodrigo       & 81.7 & 78.6 & 85.9 & 78.6 & 61.1 & \colorbox{gray!30}{2.6} & 95.9 \\
        & ICFHR\(_{2016}\)  & 74.7 & 76.5 & 75.3 & 74.1 & 75.9 & 74.9 & \colorbox{gray!30}{7.8} \\ \midrule

        \multirow{7}{*}{Michael \cite{evaluating_michael_2019}}
        & IAM           & \colorbox{gray!30}{7.5} & 35.5 & 43.6 & 43.5 & 55.3 & 65.3 & 85.2 \\
        & Rimes         & 54.5 & \colorbox{gray!30}{6.9} & 63.9 & 70.2 & 64.3 & 66.8 & 86.3 \\
        & G.W.          & 78.9 & 80.5 & \colorbox{gray!30}{53.8} & 73.4 & 79.5 & 76.7 & 100.0 \\
        & Bentham       & 49.1 & 60.0 & 50.7 & \colorbox{gray!30}{8.5} & 74.5 & 70.7 & 91.2 \\
        & S.G.          & 100.0 & 100.0 & 100.0 & 100.0 & \colorbox{gray!30}{76.9} & 79.5 & 100.0 \\
        & Rodrigo       & 100.0 & 100.0 & 87.2 & 100.0 & 92.9 & \colorbox{gray!30}{3.8} & 92.4 \\
        & ICFHR\(_{2016}\)  & 89.3 & 89.7 & 81.6 & 80.4 & 77.9 & 77.9 & \colorbox{gray!30}{9.5} \\ \midrule

        \multirow{7}{*}{LT \cite{light_barrere_2022}}
        & IAM           & \colorbox{gray!30}{7.9} & 30.8 & 32.3 & 33.8 & 51.4 & 48.4 & 98.1 \\
        & Rimes         & 42.9 & \colorbox{gray!30}{5.0} & 52.5 & 61.9 & 60.8 & 56.4 & 90.5 \\
        & G.W.          & 83.5 & 85.8 & \colorbox{gray!30}{79.6} & 87.9 & 75.0 & 73.3 & 100.0 \\
        & Bentham       & 42.0 & 55.4 & 39.6 & \colorbox{gray!30}{6.0} & 65.4 & 51.7 & 92.4 \\
        & S.G.          & 86.6 & 82.1 & 95.6 & 84.9 & \colorbox{gray!30}{12.5} & 65.9 & 96.7 \\
        & Rodrigo       & 73.1 & 67.6 & 80.9 & 70.8 & 37.8 & \colorbox{gray!30}{2.0} & 90.7 \\
        & ICFHR\(_{2016}\)  & 78.0 & 81.3 & 81.3 & 77.9 & 77.4 & 76.5 & \colorbox{gray!30}{5.9} \\ \midrule

        \multirow{7}{*}{VLT \cite{training_barrere_2024}}
        & IAM           & \colorbox{gray!30}{8.9} & 29.4 & 32.1 & 33.3 & 48.2 & 47.4 & 89.7 \\
        & Rimes         & 44.9 & \colorbox{gray!30}{5.1} & 56.4 & 63.2 & 62.4 & 59.1 & 96.4 \\
        & G.W.          & 69.5 & 72.2 & \colorbox{gray!30}{25.2} & 69.6 & 76.3 & 75.9 & 100.0 \\
        & Bentham       & 41.3 & 53.7 & 43.7 & \colorbox{gray!30}{6.1} & 65.8 & 53.5 & 85.1 \\
        & S.G.          & 91.8 & 83.3 & 98.4 & 86.9 & \colorbox{gray!30}{9.2} & 58.9 & 100.0 \\
        & Rodrigo       & 71.7 & 67.2 & 80.2 & 72.0 & 38.7 & \colorbox{gray!30}{2.2} & 90.7 \\
        & ICFHR\(_{2016}\)  & 78.3 & 80.3 & 80.8 & 81.7 & 79.5 & 81.8 & \colorbox{gray!30}{6.0} \\
        \bottomrule
    \end{tabular}
\end{table}

\begin{table}[ht!]
    \caption{Complete CER results in all datasets using synthetic data.}
    \label{tab:all_results_synth}
    \renewcommand{\arraystretch}{1.3} 
    \centering
    \tiny
    \setlength{\tabcolsep}{1.6pt} 
    \begin{tabular}{llccccccl}
        \toprule
        Method & S/T & IAM & Rimes & G.W. & Bentham & S.G. & Rodrigo & ICFHR$_{2016}$ \\ \midrule
        \multirow{5}{*}{CRNN \cite{multidimensional_recurrent_puig_2017}} 
        & WIT-en & 11.9 & 22.3 & 16.9 & 26.9 & 25.8 & 28.1 & 78.5 \\
        & WIT-fr & 19.5 & 17.0 & 26.3 & 31.7 & 26.3 & 27.2 & 78.1 \\
        & WIT-es & 20.0 & 22.7 & 27.4 & 32.9 & 27.2 & 22.2 & 79.4 \\
        & WIT-la & 20.7 & 24.6 & 27.4 & 34.5 & 21.4 & 27.4 & 79.0 \\
        & WIT-de & 20.4 & 25.3 & 27.3 & 32.1 & 28.8 & 28.3 & 77.6 \\ \midrule

        \multirow{5}{*}{VAN \cite{endtoend_coquenet_2022}}
        & WIT-en & 16.9 & 24.3 & 25.8 & 26.1 & 26.3 & 28.4 & 75.0 \\
        & WIT-fr & 22.4 & 19.4 & 31.7 & 33.1 & 25.3 & 25.8 & 76.9 \\
        & WIT-es & 23.1 & 23.5 & 32.6 & 34.8 & 26.0 & 23.2 & 77.5 \\
        & WIT-la & 22.7 & 24.5 & 34.1 & 34.8 & 23.5 & 28.2 & 77.7 \\
        & WIT-de & 21.9 & 26.0 & 30.8 & 33.8 & 28.2 & 29.6 & 74.3 \\ \midrule

        \multirow{5}{*}{C-SAN \cite{arce_self-attention_2022}}
        & WIT-en & 32.0 & 38.0 & 47.8 & 42.7 & 35.7 & 38.8 & 83.5 \\
        & WIT-fr & 35.7 & 35.9 & 47.4 & 46.5 & 36.4 & 40.4 & 81.5 \\
        & WIT-es & 36.2 & 38.0 & 46.4 & 47.4 & 35.0 & 37.3 & 82.7 \\
        & WIT-la & 36.6 & 38.8 & 49.6 & 48.0 & 35.8 & 40.7 & 83.4 \\
        & WIT-de & 34.8 & 38.6 & 48.9 & 46.0 & 36.6 & 40.5 & 81.1 \\ \midrule

        \multirow{5}{*}{HTR-VT \cite{li_htr-vt_2025}}
        & WIT-en & 20.7 & 31.5 & 26.3 & 28.3 & 29.4 & 32.2 & 77.5 \\
        & WIT-fr & 27.5 & 26.6 & 34.2 & 38.1 & 29.3 & 32.1 & 78.1 \\
        & WIT-es & 28.2 & 30.3 & 35.2 & 38.6 & 30.1 & 26.6 & 77.3 \\
        & WIT-la & 28.9 & 33.1 & 35.4 & 39.6 & 27.8 & 30.6 & 78.2 \\
        & WIT-de & 29.2 & 34.1 & 36.6 & 39.2 & 33.0 & 34.2 & 76.7 \\ \midrule

        \multirow{5}{*}{Kang \cite{pay_attention_kang_2022}}
        & WIT-en & 28.7 & 44.5 & 45.1 & 51.3 & 40.0 & 46.1 & 85.3 \\
        & WIT-fr & 38.6 & 33.1 & 47.7 & 57.4 & 36.2 & 41.9 & 87.8 \\
        & WIT-es & 37.7 & 49.1 & 70.3 & 66.4 & 42.8 & 48.7 & 100.0 \\
        & WIT-la & 26.2 & 29.6 & 39.6 & 41.2 & 22.7 & 35.3 & 95.3 \\
        & WIT-de & 32.3 & 40.1 & 42.0 & 43.9 & 32.4 & 43.0 & 84.2 \\ \midrule

        \multirow{5}{*}{Michael \cite{evaluating_michael_2019}}
        & WIT-en & 20.6 & 34.0 & 30.5 & 36.6 & 35.2 & 43.4 & 83.6 \\
        & WIT-fr & 32.9 & 25.2 & 42.8 & 52.4 & 36.8 & 41.1 & 83.5 \\
        & WIT-es & 33.8 & 33.9 & 45.0 & 54.0 & 36.6 & 33.4 & 85.2 \\
        & WIT-la & 37.7 & 39.5 & 47.3 & 58.5 & 30.7 & 45.6 & 82.9 \\
        & WIT-de & 39.1 & 44.1 & 48.4 & 61.7 & 40.3 & 49.7 & 79.8 \\ \midrule

        \multirow{5}{*}{LT \cite{light_barrere_2022}}
        & WIT-en & 13.8 & 25.1 & 16.3 & 21.4 & 23.4 & 28.0 & 80.6 \\
        & WIT-fr & 22.5 & 18.9 & 26.2 & 35.5 & 24.1 & 27.7 & 79.8 \\
        & WIT-es & 23.3 & 25.0 & 27.1 & 38.0 & 22.5 & 24.8 & 81.4 \\
        & WIT-la & 24.0 & 26.4 & 27.9 & 39.3 & 22.5 & 28.5 & 83.0 \\
        & WIT-de & 23.1 & 27.6 & 24.9 & 36.3 & 26.0 & 28.2 & 78.8 \\ \midrule

        \multirow{5}{*}{VLT \cite{training_barrere_2024}}
        & WIT-en & 15.3 & 26.7 & 19.0 & 25.0 & 23.5 & 29.1 & 80.6 \\
        & WIT-fr & 23.0 & 19.5 & 25.9 & 37.3 & 23.3 & 27.9 & 79.5 \\
        & WIT-es & 24.9 & 27.0 & 28.4 & 40.6 & 25.1 & 24.2 & 79.8 \\
        & WIT-la & 26.0 & 28.3 & 29.9 & 42.0 & 26.6 & 31.0 & 82.1 \\
        & WIT-de & 23.6 & 25.9 & 27.0 & 39.2 & 24.4 & 28.2 & 79.2 \\
        \bottomrule
    \end{tabular}
\end{table}



\begin{table}[h]
\caption{Zero-shot performance (CER) of VLMs on HTR datasets vs. the best-reported OOD results in the paper (HTR$_\text{OOD}$ column).}
\label{tab:vlm_comparison}
    \small
    \centering
    \renewcommand{\arraystretch}{1.2} 
    \setlength{\tabcolsep}{1.1pt}
   \begin{tabular}{@{}lccccc@{}}
    \toprule
    \textbf{Dataset} & LLaVA1.6 & Kosmos-2 & TrOCR$_{\text{M}}$ & InstructBlip & HTR$_\text{OOD}$ \\ \midrule
        IAM & 74.9 & 80.3 & 6.8 & 78.9 & 28.6 \\
        Rimes & 93.5 & 81.5 & 27.2 & 80.4 & 21.3 \\
        G.W. & 78.6 & 79.7 & 17.3 & 83.3 & 31.1 \\
        S.G. & 80.4 & 82.5 & 44.1 & 87.5 & 25.3 \\
        Bentham & 85.4 & 78.4 & 17.9 & 76.7 & 33.6 \\
        Rodrigo & 76.4 & 81.2 & 38.1 & 86.2 & 38.5 \\
        ICFHR$_{2016}$ & 95.3 & 87.2 & 92.6 & 88.7 & 75.3 \\ \bottomrule
    \end{tabular}
    \end{table}


\section{Hyperparameters}
\label{sec:A_hyperparameters}

\subsection{Architectures implementation}
As stated in the main paper, we aimed to follow the implementation closest to the original papers using the available information for those that did not provide code. In all cases, the most significant architectural change occurred in the final prediction layer, where the output vocabulary size was adjusted to match the vocabulary size (94) reported in Section \ref{sec:methodology} of the main text. 

\subsection{Data augmentation}
We detail the parameters used for data augmentation during training. No transformations are applied during validation or testing, except for padding, which is applied equally across the validation, training, and test splits. All transformations are applied independently with a 50\% probability. For the transformations, we utilized those available in version 2 of transformations in torchvision (\texttt{torchvision.transforms.v2}). To simplify visualization and shorten the names, we directly referenced the v2 submodule. 
For operations involving OpenCV, we employed the opencv-python library (\texttt{cv2} module) to execute OpenCV transformations directly.

\begin{itemize}
    \item \textbf{Dilation} (\texttt{Custom transformation}):
    \begin{itemize}
        \item Parameters: kernel size = 3; iterations = 1.
    \end{itemize}

    \item \textbf{Erosion} (\texttt{Custom transformation}):
    \begin{itemize}
        \item Parameters: kernel size = 2; iterations = 1.
    \end{itemize}

    \item \textbf{Elastic Transform} (\texttt{v2.ElasticTransform}):
    \begin{itemize}
        \item Parameters: sigma = 5.0; alpha = 5.0; fill = 255 (white).
    \end{itemize}

    \item \textbf{Random Affine (Rotation, Translation, Shear)} (\texttt{v2.RandomAffine}):
    \begin{itemize}
        \item Parameters: rotation degrees = $\pm1$; translation = 1\% horizontally and up to 5\% vertically; shear = $\pm1$ pixels (sheared by a factor of 5); fill = 255 (white).
    \end{itemize}

    \item \textbf{Perspective} (\texttt{v2.RandomPerspective}):
    \begin{itemize}
        \item Parameters: Distortion scale = 0.1; fixed probability of applying the distortion = 100\%; fill = 255 (white).
    \end{itemize}

    \item \textbf{Gaussian Blur (Noise)} (\texttt{v2.GaussianBlur}):
    \begin{itemize}
        \item Parameters: kernel size = 3; sigma = 2.0.
    \end{itemize}

    \item \textbf{Padding} (\texttt{v2.Pad}):
    \begin{itemize}
        \item Parameters: padding = 15 pixels on the left and right; fill = 255 (white).
    \end{itemize}

    \item \textbf{Grayscale} (\texttt{v2.Grayscale}):
    \begin{itemize}
        \item  Parameters: \text{num\_output\_channels = 1}
    \end{itemize}

    \item \textbf{Convert to Tensor} (\texttt{v2.ToTensor}):
    \begin{itemize}
        \item Converts the input data to a PyTorch tensor format.
    \end{itemize}
\end{itemize}

\paragraph{Dilation details.} The image is first inverted using \texttt{cv2.bitwise\_not}. Then, \texttt{cv2.dilate} is applied with the selected kernel, expanding the white areas in the image. The process is repeated for the specified number of iterations. Finally, the image is inverted again to restore its original colors.

\paragraph{Erosion details.} The image is inverted using \texttt{cv2.bitwise\_not}, followed by \texttt{cv2.erode} with the selected kernel, shrinking the white areas. This operation is also repeated for the given number of iterations. Afterward, the image is inverted back to its original color scheme.

\section{Visual and textual divergences}
\label{sec:C_metrics_fa}
In this section, we present the specific numerical metrics for visual and textual divergence across the various domains used in the factor analysis. Prior to presenting these results, we first describe the training procedure for the Convolutional Autoencoder (AE) ($\phi_{\theta_{S}}$) employed to measure reconstruction error (visual divergence).

\subsection{Convolutional Autoencoder}
\subsubsection{Architecture}
Regarding the Autoencoder (AE) used, we employed a rather simple convolutional architecture. The encoder progressively downsamples and compresses the input image into a $512$-dimensional latent vector using four $3\times3$ convolutional layers, each followed by leaky ReLU activation and $2\times2$ max-pooling. The feature channels increase sequentially from $1$ to $16, 32, 64$, and 128, with a fully connected layer producing the final latent representation. The decoder reconstructs the image from the latent vector by reversing the encoder's process. It uses a fully connected layer to reshape the latent vector into a tensor, followed by four transposed convolutional layers that upsample the feature map to the original image size. Feature channels decrease from $128$ to $64, 32, 16$, and finally $1$, with leaky ReLU activations applied after each layer, except the final layer, which uses a sigmoid activation to normalize output pixel values. 

Despite the simplicity of the architecture, the input images are rescaled to dimensions of 64 pixels in height and 1024 pixels in width, result in a model with approximately 33 million parameters. Note that due to the large image size, the pre-flattened vector resulting from the encoder's downsampling has 32,768 dimensions (flattening the final feature map of the encoder with 128 channels, a height of 4, and a width of 64). Using an MLP to reduce this vector to 512 dimensions significantly increases the parameter count. These two layers (one in the encoder and one in the decoder) account for 99\% of the model's parameters.

\subsubsection{Training details}
We train the AE to minimize the Mean Squared Error (MSE) between the input and reconstructed images. We employ the Adam optimizer with a learning rate of 0.001 for a maximum of 100 epochs. To avoid overfitting, we save the best-performing model according to the validation loss of the same source domain at the end of each epoch. 

\subsection{Visual divergence}
\begin{figure}[h]
    \centering
    \includegraphics[width=1.0\linewidth]{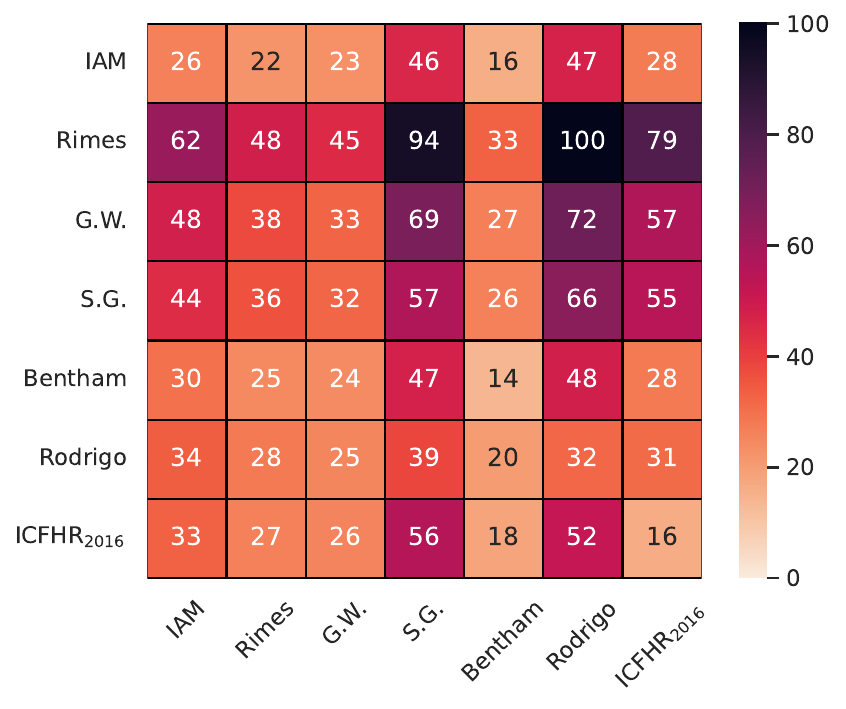}
    \caption{Heatmap of visual divergence between source (rows) and target (columns) from real HTR domains. Divergence values are normalized, with higher scores indicating greater divergence and lower scores reflecting lower divergence.}
    \label{fig:rec_errors}
\end{figure}

\begin{figure*}[h]
    \centering
\includegraphics[width=1.0\linewidth]{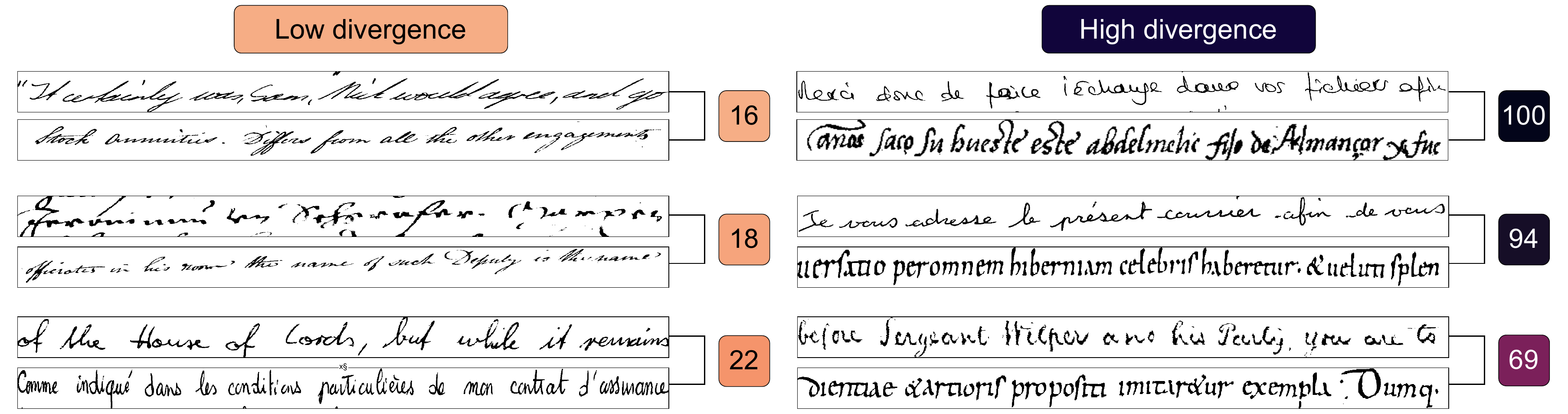}
    \caption{Representation of visual divergence between domains. Examples are presented in two columns: on the left, three domain pairs with low visual divergence, and on the right, three pairs with high visual divergence. For each domain pair, the source domain is shown at the top and the target domain at the bottom. Left (top to bottom): First pair (IAM-Bentham), second pair (ICFHR$_{2016}$-Bentham), third pair (IAM-Rimes). Right (top to bottom): First pair (Rimes-Rodrigo), second pair (Rimes-S.G.), third pair (G.W.-S.G.). The divergence percentage (see Fig. \ref{fig:rec_errors}) is displayed for each pair.}
    \label{fig:visual_divergences_examples}
\end{figure*}

This section presents the results of visual domain divergence, measured by the reconstruction error obtained from the autoencoder described in previous sections. Fig. \ref{fig:rec_errors} illustrates the divergence (calculated as the average MSE per image) between each domain pair, with the source represented on the Y-axis and target on the X-axis. Divergences are computed between training and test splits for each pair. To facilitate interpretability, the values are normalized, such that a value of 100 reflects high divergence (darker colors), while a value of 0 denotes indicating low divergence (lighter colors). To validate the visual divergence results in the OOD scenario, Fig. \ref{fig:visual_divergences_examples} presents images from three pairs of domains with low visual divergence (left) and three domains with high one (right) with their respective scores. Note that the left column features writing styles with very similar stroke densities, while the right column displays styles that differ significantly in both stroke appearance and density. The domain pairs were selected based on the scores presented in Fig. \ref{fig:rec_errors}, ensuring minimal repetition of domains to better highlight the differences.

\begin{figure}[h]
    \centering
    \includegraphics[width=1.0\linewidth]{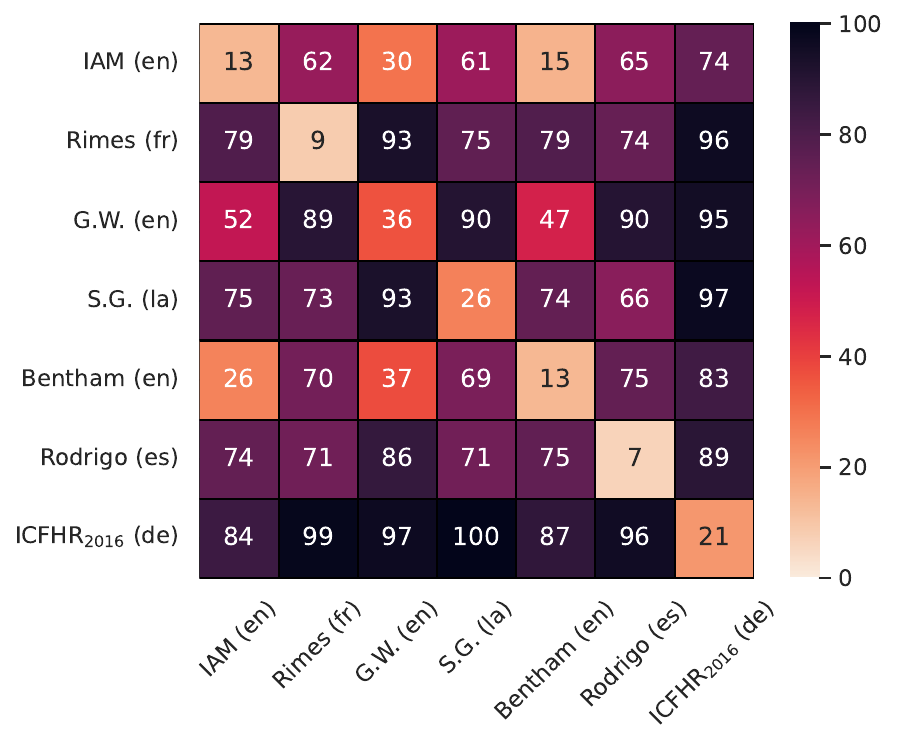}
    \caption{Heatmap of textual divergence from real HTR domains. Rows correspond to source domains, while columns represent target domains. The values are normalized, with 100 indicating maximum divergence and 0 representing minimum divergence. 
    }
    \label{fig:kl_div_real}
\end{figure}

\begin{figure}[h]
    \centering
    \includegraphics[width=1.0\linewidth]{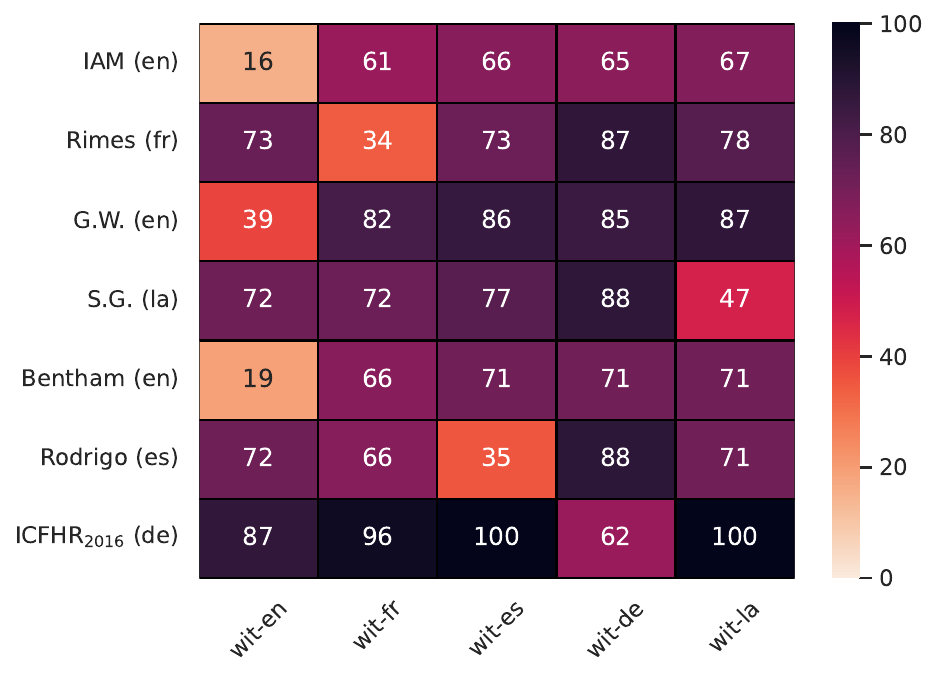}
    \caption{Heatmap of textual divergence between real and synthetic domains. 
    The values are normalized, where 100 represents maximum divergence and 0 represents minimum divergence. Note that each source domain corresponds to a target domain that matches its language, except for English, where three target domains are used: IAM, George Washington, and Bentham.}
    \label{fig:kl_div_synth}
\end{figure}

\subsection{Textual divergence}
We present the results of textual domain divergences, quantified as the averaged KL-divergence across n-grams as described in the main text. Fig. ~\ref{fig:kl_div_real} shows the divergence between textual distributions across domains, with the source represented on the Y-axis and the target the X-axis.  Divergences are computed between training and test splits for each pair. Fig.~\ref{fig:kl_div_synth} presents the textual divergences between real source domains (Y-axis) and synthetic domains (X-axis) for each language. In this case, the divergence is calculated between the training split of the source domain and the training split of the synthetic data, as these are used to compute the n-grams and train the models in the synthetic experiments. Both figures display normalized values, where a value of 100 indicates maximum divergence (darker colors) and 0 minimum divergence between texts.


\section{Factor analysis}
\label{sec:D_factor_analysis}
\begin{figure}[ht]
    \centering
    \includegraphics[width=0.53\columnwidth]{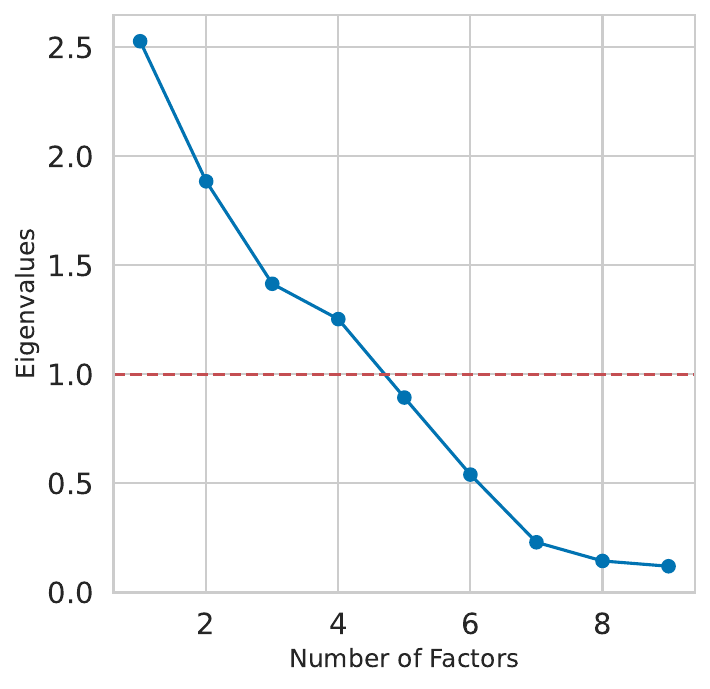}
    \caption{Scree plot: Eigenvalues of the standardized values used for factor analysis, ordered in descending magnitude. We chose to retain 4 factors, as these are the ones with eigenvalues $\geq1$.}
    \label{fig:scree_plot_fa}
\end{figure}

The selection of the number of factors is a crucial criterion for analyzing the outcomes of the factor analysis. Note that the first $k$ factors span the subspace defined by the first $k$ eigenvectors of the data matrix. To determine the number of factors $(n)$, the simplest rule of thumb involves retaining all eigenvectors with eigenvalues $\geq 1$. This can be simply visualized by plotting the eigenvalues in descending order using a scree plot, as shown in Fig. \ref{fig:scree_plot_fa}. Based on this analysis, we decided to retain four factors as stated in the main text of the paper.

\end{document}